\pdfoutput=1
\documentclass[journal]{IEEEtran}

\usepackage{amssymb}
\usepackage{amsmath}
\usepackage{booktabs}
\usepackage{algorithm}
\usepackage{mathtools}
\usepackage{verbatim}
\usepackage{url}
\usepackage{caption}
\usepackage{graphicx}
\usepackage{epstopdf}
\usepackage{mathtools}
\usepackage{stmaryrd}
\usepackage{subfloat}
\usepackage[caption=false]{subfig}
\usepackage[noend]{algpseudocode}
\usepackage[mathscr]{eucal}
\usepackage{cleveref}

\usepackage[dvipsnames]{xcolor}
\algdef{SE}[DOWHILE]{Do}{doWhile}{\algorithmicdo}[1]{\algorithmicwhile\ #1}%

\begin{document}
\title{Attn-HybridNet: Improving Discriminability of Hybrid Features with Attention Fusion}

\author{Sunny~Verma,
        Chen~Wang,
        Liming~Zhu,~\IEEEmembership{Member,~IEEE}
        and~Wei~Liu,~\IEEEmembership{Senior Member,~IEEE}
\thanks{Sunny Verma is with The Data Science Institute, University of Technology Sydney, Australia (email: Sunny.Verma@uts.edu.au}
\thanks{Wei Liu is with the School of Computer Science, University of Technology Sydney, Australia (email: Wei.Liu@uts.edu.au).}
\thanks{Chen Wang and Liming Zhu are with Data61, Commonwealth Scientific and Industrial Research Organization, CSIRO, Sydney, Australia (email: Chen.Wang@data61.csiro.au, Liming.Zhu@data61.csiro.au).}
\thanks{The source code of proposed technique and extracted features are available at https://github.com/sverma88/Attn-HybridNet---IEEE-TCYB.}
}
\markboth{}%
{Verma \MakeLowercase{\textit{et al.}}: Attn-HybridNet}

\maketitle

\begin{abstract}
The principal component analysis network (PCANet) is an unsupervised parsimonious deep network, utilizing principal components as filters in its convolution layers. Albeit powerful, the PCANet consists of basic operations such as \textit{principal components} and spatial pooling, which suffers from two fundamental problems. First, the \textit{principal components} obtain information by transforming it to column vectors  (which we call the amalgamated view), which incurs the loss of the spatial information in the data. Second, the generalized spatial pooling utilized in the PCANet induces feature redundancy and also fails to accommodate spatial statistics of natural images. In this research, we first propose a tensor-factorization based deep network called the Tensor Factorization Network (TFNet). The TFNet extracts features from the spatial structure of the data (which we call the minutiae view). We then show that the information obtained by the PCANet and the TFNet are distinctive and non-trivial but individually insufficient. This phenomenon necessitates the development of proposed \textit{HybridNet}, which integrates the information discovery with the two views of the data. To enhance the discriminability of hybrid features, we propose \textit{Attn-HybridNet}, which alleviates the feature redundancy by performing attention-based feature fusion. The significance of our proposed \textit{Attn-HybridNet} is demonstrated on multiple real-world datasets where the features obtained with \textit{Attn-HybridNet} achieves better classification performance over other popular baseline methods, demonstrating the effectiveness of the proposed technique.
\end{abstract}


\IEEEpeerreviewmaketitle

\section{Introduction}

Feature engineering is an essential task in the development of machine learning systems and has been well-studied with substantial efforts from communities including computer vision, data mining, and signal processing \cite{zheng2018sift}. In the era of deep learning, the features are extracted by processing the data through multiple stacked layers in deep neural networks. These deep neural networks sequentially perform sophisticated operations to discover critical information concealed in the data \cite{bengio2013representation}. However, the training time required to obtain these superior data representations is exponentially large as these networks have an exhaustive hyper-parameter search space and usually suffer from various training difficulties \cite{glorot2010understanding}. Besides, the deep networks are complex models that require high computational resources for their training and deployment. Hence, this limits the usability of these networks on micro-devices such as cellphones \cite{passalis2018training,han2015deep}. The current research trend focuses on alleviating the memory and space requirements associated with the deep networks \cite{kossaifi2017tensor}.

In this regard, to produce lightweight convolution neural networks (CNNs) architecture, most of the existing solutions would 1) approximate the convolution and fully connected layers by factorization \cite{zhang2015efficient,lebedev2014speeding}, 2) compress the layers with quantization/hashing \cite{han2015deep,cheng2017quantized}, or, 3) replace the fully connected layer with a tensorized layer and optimize the weights of this layer by retraining \cite{kossaifi2017tensor}. However, all these techniques require a pre-trained CNN network and can only work as post-optimization technique. By contrast, the goal of this research is to build lightweight deep networks which are computationally inexpensive to train. In other words, this research aims to build deep networks that are independent of a) high-performance hardware and b) exhaustive hyper-parameter search space, where a PCANet represents one such promising architecture.

The PCANet is an unsupervised deep parsimonious network utilizing \emph{principal components} as convolution filters for extracting features in its cascaded layers \cite{chan2015pcanet}. Due to the remarkable performance of PCANet on several benchmark face datasets, the network is recognized as a simple deep learning baseline for image classification. However, the features extracted by PCANet do not achieve competitive performance on challenging object recognition datasets such as CIFAR-10 \cite{krizhevsky2009learning}. There are two major reasons for this performance limitation: 1) the PCANet vectorizes the data while extracting \emph{principal components} which results in loss of spatial information exhibiting in the data and, 2) the output layer (which is spatial-pooling) utilized in the PCANet induces feature redundancy and does not adapt to the structure of natural images, deteriorating the performance of classifiers \cite{jia2013compact,jia2012beyond}. However, the vectorization of the data is inherent with the \emph{principal components} and hence motivates to devise techniques that can alleviate the loss of spatial information present in the data. In other words, techniques that can extract information from the untransformed view of the data\footnote{Throughout this paper we refer to the vectorized presentation of the data as the amalgamated view where all modes of the data (also called dimension for higher order-matrices, i.e. tensors) are collapsed to obtain a vector. The untransformed view of the data, i.e., when viewed with its multiple modes (e.g., tensors), is referred to as the minutiae view of the data.} which is proven to be beneficial in literature \cite{vasilescu2002multilinear,verma2017extracting,chien2017tensor}. 



In this research, we first propose an unsupervised tensor factorization based deep network called {Tensor Factorization Network} (TFNet).  \emph{The TFNet, contrary to the PCANet, does not vectorize the data while obtaining weights for its convolution filters.}{} Therefore, it is able to extract information associated with the spatial structure of the data or the minutiae view of the data. Besides, the information is independently obtained from each mode of the data, providing several degrees of freedom to the information extraction procedure of TFNet.

Importantly, we hypothesize that the information obtained from either the amalgamated view or the minutiae view of the data is essential but individually insufficient as they respectively conceal complementary information associated with the two views\footnote{Throughout this paper by two views, we mean the amalgamated view and the minutiae view.} of the data \cite{liu2013mining,verma2017extracting}. Therefore, the integration of information from these two views can enhance the performance of classification systems. To this end, we propose the \textit{Hybrid Network} (\textit{HybridNet}) that integrates information discovery and feature extraction from the minutiae view and the amalgamated view of the data simultaneously in its consolidated architecture.

Although the \textit{HybridNet} reduces the information loss by integrating information from the two views of the data, it may still suffer from feature redundancy problems arising from the generalized spatial pooling utilized in the output layer of PCANet. Therefore, we propose an attention-based fusion scheme \textit{Attn-HybridNet} that performs feature selection and aggregation, thus enhancing the feature discriminability in hybrid features. 

The superiority of feature representations obtained with the \textit{Attn-HybridNet} is validated by performing comprehensive experiments on multiple real-world benchmark datasets. The differences and similarities between the PCANet, TFNet, \textit{HybridNet}, and \textit{Attn-HybridNet} from data view perspectives are summarized in \Cref{compmethod}.

Our contributions in this paper are summarized below:
\begin{itemize}

\item We propose \textit{Tensor Factorized Network} (TFNet), which extracts features from the minutiae view of the data and hence is able able to preserve the spatial information present in the data that is proven beneficial for image classification.

\item We propose Left one Mode Out Orthogonal Iteration (\textit{LoMOI}) algorithm, which optimizes convolution weights from the minutiae view of the data utilized in the proposed TFNet.

\item We introduce the \textit{Hybrid Network} (\textit{HybridNet}), which integrates the feature extraction and information discovery procedure from two views of the data. This integration procedure reduces information loss from the data by combining the merits of the PCANet and TFNet and obtains superior features from both of the two schemes.

\item We propose the \textit{Attn-HybridNet}, which alleviates feature redundancy among hybrid features by performing feature selection and aggregation with an attention-based fusion scheme. The \textit{Attn-HybridNet} enhances the discriminability of the feature representations, which further theoretically improves the classification performance of our scheme.

\item {We perform comprehensive evaluations and case studies to demonstrate the effectiveness of features obtained by \textit{Attn-HybridNet} and \textit{HybridNet} on multiple benchmark real-world datasets.}

\end{itemize}

The rest of the paper is organized as follows: in Sec.~\ref{sec:bg} we present the literature review including prior works and background on PCANet and tensor preliminaries. We then present the details of our proposed TFNet, \textit{HybridNet}, and \textit{Attn-HybridNet} Sec.~\ref{sec:TD}, Sec.~\ref{sec:HD}, and Sec.~\ref{prop_attn} respectively. Next we describe our experimental setup, results and discussions in Sec.~\ref{sec:exp} and Sec.~\ref{ressec}. Finally, we conclude our work and specify the future directions for its improvement in Sec.~\ref{sec:con}.

\begin{table}[t]
\scriptsize
\captionsetup{justification=centering}
\centering
\begin{tabular}{@{}l|ccc@{}}
\toprule[1pt]
\begin{tabular}[c]{@{}l@{}} Methods \end{tabular} & \multicolumn{1}{c}{\begin{tabular}[c]{@{}c@{}}Amalgamated View \end{tabular}} & \multicolumn{1}{c}{\begin{tabular}[c]{@{}c@{}}Minutiae View  \end{tabular}}  &  \multicolumn{1}{c}{\begin{tabular}[c]{@{}c@{}} Attention Fusion \end{tabular}}  \\ \midrule
PCANet \cite{chan2015pcanet}     &   \checkmark   &    $\times$  &  $\times$  \\ \midrule
TFNet \cite{VermaL0Z18}     &  $\times$     &   \checkmark      & $\times$      \\ \midrule
\textit{HybridNet} \cite{VermaL0Z18}    &  \checkmark     &   \checkmark     &  $\times$      \\ \midrule
\textit{Attn-HybridNet}  &  \checkmark   &   \checkmark      &  \checkmark      \\ \bottomrule[1pt]
\end{tabular}
\caption{Comparison of different feature extraction models}
\label{compmethod}
\vspace{-5mm}
\end{table}


\section{Literature Review}
\label{sec:bg}
The success of utilizing CNNs for multiple computer vision tasks such as visual categorization, semantic segmentation, etc. has lead to drastic research and development in the deep learning field. At the same time, to supersede human performance on these tasks, the CNNs requires an enormous amount of computational resources. For example, the ResNet \cite{he2016deep} achieves a top-5 error rate of $3.57\%$ that consists of $152$-layers accounting for a total of $60$M parameters and requires $2.25 \times 10^{10}$ flops of the data at inference. This substantial computational cost restricts the applicability of such models on devices with limited computational resources such as mobile devices. Reducing the size and time complexities of the CNNs has, therefore, become a non-trivial task for their practical applications. In this regard, researchers actively pursuit three main active research directions: 1) compression of trained CNNs weights with quantization, 2) approximating convolution layer with factorization, and 3) replacing fully connected layers with custom-built layers. 

In the first category, the aim is to reduce the size of trained CNNs by compressing its layers with quantization or hashing, as in \cite{han2015deep,cheng2017quantized,chen2015compressing}. These quantized CNN models achieve similar recognition accuracy with significantly less requirements for computational resources during inference. Similarly, the works in \cite{zhang2015efficient,lebedev2014speeding} obtain approximations of fully connected and convolution layers by utilizing factorization for compressing the CNN models. However, both the quantization and factorization based methods compress a pre-trained CNN model instead of building a smaller or faster CNN model in the first place. Therefore, these techniques inherit the limitations of the pre-trained CNN models. 

In the second and the third categories, the aim is to replace fully connected layers by customized lightweight layers that substantially reduce the size of any CNN model. For example, in \cite{kossaifi2017tensor} proposes a neural tensor layer while the work in \cite{passalis2018training} proposes a BoF (Bag-of-features) model as a neural pooling layer. These techniques augment conventional CNNs layers and produce their lightweight versions which are trainable in an end-to-end fashion. However, a major limitation of these work is that they are only capable of replacing a fully connected layer, and in order to replace a convolution layer, they usually end up functioning similarly to the work in the first category.

Different from the above research, possible solutions for obtaining lightweight CNN architecture with lower computational requirements on smaller size images are proposed in PCANet \cite{chan2015pcanet} and TFNN \cite{chien2017tensor}. The PCANet is a deep unsupervised parsimonious feature extractor, whereas TFNN is a supervised CNN architecture utilizing neural tensor factorizations for extracting information from multiway data. Both these networks achieve very high classification performance on handwritten digits dataset but fail to obtain competitive performance on object recognition dataset. This is because the PCANet (and its later variants FANet \cite{huang2015fanet}) incur information loss associated with the spatial structure of the data as it obtains weights of its convolution filters from the amalgamated view of the data. Contrarily, the TFNN extracts information by isolating each view of the multi-view data and fails to efficiently consolidate them for their utmost utilization, incurring the loss of common information present in the data.    

Therefore, the information from both the amalgamated view and the minutiae view is essential for classification, and their integration can enhance the classification performance \cite{liu2013mining,verma2017extracting}. In this research, we first propose \textit{HybridNet}, which integrates the two kinds of information in its deep parsimonious feature extraction architecture. A major difference between \textit{HybridNet} and PCANet is that the \textit{HybridNet} obtains information from both views of the data simultaneously, whereas the PCANet is restricted to obtain information from the amalgamated view of the data. The \textit{HybridNet} is also notably different from TFNN as the \textit{HybridNet} is an unsupervised deep network while the TFNN is a supervised deep neural network. Moreover, the \textit{HybridNet} extracts information from minutiae view of the data, whereas the TFNN extracts information by isolating each mode of multi-view data. 

Moreover, to enhance the discriminability of the features obtained with \textit{HybridNet}, we propose the \textit{Attn-HybridNet}, which performs attention-based fusion on hybrid features. The \textit{Attn-HybridNet} reduces feature redundancy by performing feature selection and obtains superior feature representations for supervised classification. We present the related background preliminaries in the next subsection.

\subsection{Background}
We briefly summarize PCANet's $2$-layer architecture and provide background on tensor preliminaries in this section.

\subsubsection{The First Layer} 
The procedure begins by extracting overlapping patches of size $k_1 \times k_2$ around each pixel in the image; where patches from image $\boldsymbol{I}_i$ are denoted as $\textbf{x}_{i,1}, \textbf{x}_{i,2},...,\textbf{x}_{i,\tilde{m}\tilde{n}} \in \mathbb{R} ^{ k_1 k_2} $, $\tilde{m} = m - \lceil{\frac{k_1}{2}}\rceil$\footnote{The operator $\lceil z \rceil$ gives the smallest integer greater than or equal to $z$.} and $\tilde{n} = n - \lceil{\frac{k_2}{2}}\rceil$. Next, the obtained patches are zero-centered by subtracting the mean of the image patches and \textit{vectorized} to obtain $\boldsymbol{X}_i \in \mathbb{R}^{  k_1 k_2 \times \tilde{m}\tilde{n}}$ as the patch matrix. After repeating the same procedure for all the training images we obtain $\boldsymbol{X} \in \mathbb{R} ^{ k_1 k_2 \times N\tilde{m}\tilde{n}} $ as the final patch-matrix from which the $pca$ filters are obtained. The $PCA$ minimizes the reconstruction error with orthonormal filters known as the principal eigenvectors of $\boldsymbol{X} \boldsymbol{X}^{T}$ calculated as in Eq.~\ref{eq:pca}
\begin{equation} \label{eq:pca}
\footnotesize{
\min\limits_{\boldsymbol{V} \in\mathbb{R} ^{ k_1 k_2 \times L_1}}  \| \boldsymbol{X} - \boldsymbol{V}\boldsymbol{V}^{T} \boldsymbol{X} \|_{F}, \ s.t. \quad \boldsymbol{V}^{T}V = \boldsymbol{I}_{L_1} }
\end{equation}
where $\boldsymbol{I}_{L_1}$ is an identity matrix of size $L_1 \times L_1$ and $L_1$ is the total number of obtained filters. These convolution filters can now be expressed as:
\begin{equation} \label{eq:pcafilt}
\footnotesize{
\boldsymbol{W}_{l_{PCANet}}^{1} = mat_{k_1,k_2} (ql(\boldsymbol{X} \boldsymbol{X}^{T})) \in \mathbb{R} ^{ k_1 \times k_2}}
\end{equation}
where $mat_{k_1,k_2}(v)$ is a function that maps $ v \in \mathbb{R} ^{ k_1k_2}$ to a matrix $\boldsymbol{W} \in \mathbb{R} ^{ k_1 \times k_2}$, and $ql(\boldsymbol{X} \boldsymbol{X}^{T})$ denotes the $l$-th principal eigenvector of $\boldsymbol{X} \boldsymbol{X}^{T}$. Next, each training image $\boldsymbol{I}_i$ is convolved with the $L_1$ filters as in Eq.~\ref{eq:pcaconv}.
\begin{equation} \label{eq:pcaconv}
\footnotesize{
 \boldsymbol{I}_{i_{PCANet}}^{l}  = \boldsymbol{I}_{i} \ast \boldsymbol{W}_{l_{PCANet}}^{1}}
\end{equation}
where $\ast$ denotes the 2D convolution and $i, l$ are the image and filter indices respectively. Importantly, the boundary of image $\boldsymbol{I}_{i}$ is padded before convolution to obtain $ \boldsymbol{I}_{i_{PCANet}}^{l}$ with the same dimensions as in $ \boldsymbol{I}_{i}$. From Eq.~\ref{eq:pcaconv} a total of $N \times L_1$ images are obtained and attributed as the output from the first layer.  

\subsubsection{The Second Layer}

The methodology of the second layer is similar to the the first layer. We collect overlapping patches of size $k_1 \times k_2$ around each pixel from all input images in this layer i.e., from $ \boldsymbol{I}_{i_{PCANet}}^{l}$. Next, we vectorize and zero-centre these images patches to obtain the final patch matrix denoted as $\boldsymbol{Y} \in  \mathbb{R} ^{ k_1 k_2 \times L_1 N\tilde{m}\tilde{n}}$. This patch matrix is then utilized to obtain the convolution \textit{pca} filters in layer 2 as in Eq.~\ref{eq:pcafilt2}.
\begin{equation} \label{eq:pcafilt2}
\footnotesize
 \boldsymbol{W}_{l_{PCANet}}^{2} = mat_{k_1,k_2} (ql(\boldsymbol{Y} \boldsymbol{Y}^{T})) \in \mathbb{R} ^{ k_1 \times k_2}
\end{equation}
where $l = [1,L_2]$ denotes the number of $pca$ filters obtained in this layer. Next, the input images in this layer $\boldsymbol{I}_{i_{PCANet}}^{l}$ are convolved with the learned filters $ \boldsymbol{W}_{l_{PCANet}}^{2}$ to obtain the output from this layer in Eq.~\ref{eq:pcaconv2}. These images are then passed to the feature aggregation phase as in the next subsection.
\begin{equation} \label{eq:pcaconv2}
\footnotesize
 \boldsymbol{O}_{i_{PCANet}}^{l}  = \boldsymbol{I}_{i_{PCANet}}^l \ast \boldsymbol{W}_{l_{PCANet}}^{2}
\end{equation}

\subsubsection{The Output Layer} \label{ol}
The output layer combines the output from all the convolution layers of PCANet to obtain the feature vectors. The process initiates by first binarizing each of the real-valued outputs from Eq.~\ref{eq:pcaconv2} by utilizing a Heaviside function $H( \boldsymbol{O}_{i_{PCANet}}^{l})$ on them, which converts the positive entries to $1$ otherwise $0$. Then, these $L_2$ outputs are assembled into $L_1$ batches, where all images in a batch belong to the same convolution filter in the first layer. Then, these images are combined to form a single image by applying weighted sum as in Eq.~\ref{eq:hashpca} whose pixel value is in the range $[ 0, 2^{L_2} -1]$:
\begin{equation} \label{eq:hashpca}
\footnotesize
 \mathcal{\boldsymbol{I}}_{i_{PCANet}}^{l}  = \sum_{l=1}^{L_2} 2^{l-1}H(\boldsymbol{O}_{l_{PCANet}}^2) 
\end{equation}
Next, these binarized images are partitioned into $B$ blocks and a histogram with $2^{L_2}$ bins is obtained. Finally, the histograms from all the $B$ blocks are concatenated to form a feature vector from the amalgamted view of the images in Eq.~\ref{eq:hashpcaout}. 
\begin{equation} \label{eq:hashpcaout}
\footnotesize
 f_{i_{PCANet}}  =[Bhist( \mathcal{\boldsymbol{I}}_{i_{PCANet}}^{1}),..., Bhist( \mathcal{\boldsymbol{I}}_{i_{PCANet}}^{L_1})]^T \in \mathbb{R}^{(2^{L_2})L_1B}
\end{equation}
This block-wise encoding process encapsulates the $L_1$ images from Eq.~\ref{eq:hashpca} into a single feature vector which can be utilized for any machine learning task like clustering or classification.


\subsection{Tensor Preliminaries} 
\label{sec:Lit}
Tensors are simply multi-mode arrays or higher-order\footnote{Also known as modes (dimensions) of a tensor and are analogous to rows and columns of a matrix.} matrices of dimension $>2$. In this paper, the vectors are denoted as $\boldsymbol{x}$ are called first-order tensors, whereas the matrices are denoted as $\boldsymbol{X}$ are called second-order tensors. Analogously, matrices of order-$3$ or higher are called tensors and are denoted as $\boldsymbol{\mathscr{X}}$. A few important multilinear algebraic operations utilized in this paper are described below. 

\begin{algorithm}[t]    
\scriptsize
\captionsetup{justification=justified, skip=5pt}
\caption{\textbf{Left One Mode Out Orthogonal Iteration, \textit{LoMOI}}}
\begin{algorithmic}[1]
\State\textbf {Input:} $n$-mode tensor $\boldsymbol{\mathscr{X}} \in \mathbb{R}^{i_1,i_2,...,i_n}$; factorization ranks for each mode of the tensor $[r_1...r_{m-1},r_{m+1}...r_n]$, where $ r_k \leq i_k \forall \ k \in 1,2,...,n $ and $k \neq m$; factorization error-tolerance $ \varepsilon$, and Maximum allowable iterations $= Maxiter $, $m$ = mode to discard while factorizing

\For {$i =1,2,...,n$ and $i \neq m$}    
\State $ \boldsymbol{X}_{i} \gets $ unfold tensor $\boldsymbol{\mathscr{X}}$ on mode-$i$
\State $ \boldsymbol{U}^{(i)} \gets r_i \ \text{left singular vectors of }\boldsymbol{X}_{i}$ \Comment{extract leading $r_i$ matrix factors}

\EndFor

\State $ \boldsymbol{\mathscr{G}} \gets \boldsymbol{\mathscr{X}} \times_1 (\boldsymbol{U}^{(1)}){^T} ... \times_{m-1} (\boldsymbol{U}^{(m-1)}){^T}  \times_{m+1} (\boldsymbol{U}^{(m+1)}){^T} ... \times_n (\boldsymbol{U}^{(n)}){^T}$ \Comment{Core tensor}

\State $ \hat{\boldsymbol{\mathscr{X}}} \gets  \boldsymbol{\mathscr{G}} \times_1 \boldsymbol{U}^{(1)} ... \times_{m-1} \boldsymbol{U}^{(m-1)}  \times_{m+1} \boldsymbol{U}^{(m+1)} \times_n \boldsymbol{U}^{(n)}$ \Comment{reconstructed tensor obtained by multilinear product of the core-tensor with the factor-matrices; Eq.~\ref{eq:tucker}.}

\State $loss \gets  \|  \boldsymbol{\mathscr{X}} -  \hat{\boldsymbol{\mathscr{X}}} \|$ \Comment{decomposition loss}

\State $count \gets 0$
\While{$ [(loss \ge \varepsilon) \ Or \ (Maxiter \le count) ]$} \Comment{loop until convergence}

\For {$i =1,2,...,n$ and $i \neq m$}    

\State $ \boldsymbol{\mathscr{Y}} \gets \boldsymbol{\mathscr{X}} \times_1 (\boldsymbol{U}^{(1)}){^T} ...\times_{(i-1)} (\boldsymbol{U}^{(i-1)}){^T} \times_{(i+1)}  (\boldsymbol{U}^{(i+1)}){^T}...\times_n (\boldsymbol{U}^{(n)}){^T}$ \Comment{obtain the variance in mode-$i$}

\State $ \boldsymbol{Y}_{i} \gets $ unfold tensor $\boldsymbol{\mathscr{Y}}$ on mode-$i$

\State $ \boldsymbol{U}^{(i)} \gets \mathbf{r}_i \ \text{left singular vectors of }\boldsymbol{Y}_{i}$ 

\EndFor
\State $ \boldsymbol{\mathscr{G}} \gets \boldsymbol{\mathscr{X}} \times_1 (\boldsymbol{U}^{(1)}){^T} ...\times_{(m-1)} (\boldsymbol{U}^{(m-1)}){^T} \times_{(m+1)}  (\boldsymbol{U}^{(m+1)}){^T}...\times_n (\boldsymbol{U}^{(n)}){^T}$ 

\State $ \hat{\boldsymbol{\mathscr{X}}} \gets  \boldsymbol{\mathscr{G}} \times_1 \boldsymbol{U}^{(1)}...\times_{(m-1)} \boldsymbol{U}^{(m-1)} \times_{(m+1)}  \boldsymbol{U}^{(m+1)}...\times_n \boldsymbol{U}^{(n)}$ 

\State $loss \gets  \|  \boldsymbol{\mathscr{X}} -  \hat{\boldsymbol{\mathscr{X}}} \|$
\State $count \gets count + 1$

\EndWhile
\State\textbf {Output:} $\hat{\boldsymbol{\mathscr{X}}}$ the reconstructed tensor and $[\boldsymbol{U}^{(1)}...\boldsymbol{U}^{(m-1)},\boldsymbol{U}^{(m+1)}...\boldsymbol{U}^{(n)}]$ the factor matrices 

\end{algorithmic}

\label{Algo:HOSVD}
\end{algorithm}

\paragraph{Matriziation} also known as tensor unfolding, is the operation to rearrange the elements of an $n$-mode tensor $\boldsymbol{\mathscr{X}} \in \mathbb{R}^{{i_1}\times{i_2}...\times{i_N}}$ as matrix $\boldsymbol{X}_{(n)} \in \mathbb{R}^{{i_n} \times{j}}$ on the chosen mode \textit{n}, where $ j = \big( {i_1}\ ... \times {i_{n-1}} \times {i_{n+1}}...\times{i_N} \big)$. 

\paragraph{n-mode Product} the product of an $n$-mode tensor $\boldsymbol{\mathscr{X}} \in \mathbb{R}^{{i_1}...\times{i_{m-1}} \times{i_m} \times{i_{m+1}}...\times{i_n}}$ and a matrix $\boldsymbol{A} \in \mathbb{R}^{ {j \times i{_n}}}$ is denoted as $\boldsymbol{\mathscr{X}} \times_n \boldsymbol{A} $. The resultant of this product is also a tensor $\boldsymbol{\mathscr{Y}} \in \mathbb{R}^{{i_1}\times{i_2} \times {i_{n-1}} \times{j}\times{i_{n+1}}...\times{i_n}}$ which can also be expressed through matricized tensor as $\boldsymbol{Y}_{(n)} = \boldsymbol{A}\boldsymbol{X}_{(n)} $.

\paragraph{Tensor Decomposition}
Tensor decomposition is a form of generalized matrix factorization for approximating multimode tensors. The factorization an $n$-mode tensor $\boldsymbol{\mathscr{X}} \in \mathbb{R}^{{i_1}\times{i_2}...\times{i_n}}$ obtains two sub components:  $1)$ $\boldsymbol{\mathscr{G}} \in \mathbb{R}^{{r_1}\times{r_2}...\times{r_n}}$ which is a lower dimensional tensor called the \textit{core-tensor} and, $2)$ $\boldsymbol{U}^{(j)}\in \mathbb{R}^{{r_n}\times{i_n}} \forall j=[1,n]$ which are matrix factors associated with each mode of the tensor. The entries in the \textit{core-tensor} $\boldsymbol{\mathscr{G}}$ signify the interaction level between tensor elements. The factor matrices $\boldsymbol{U}^{(n)}$ are analogous to \emph{principal components} associated with the respective mode-$n$.
This scheme of tensor factorization falls under the \textit{Tucker} family of tensor decomposition \cite{cichocki2015tensor}. The original tensor  $\boldsymbol{\mathscr{X}}$ can be reconstructed by taking the n-mode product of the \textit{core-tensor} and the factor matrices as in Eq.~\ref{eq:tucker}.
\begin{equation} \label{eq:tucker}
\boldsymbol{\mathscr{G}} \times_1 \boldsymbol{U}^{(1)} \times_2 \boldsymbol{U}^{(2)}... \times_N \boldsymbol{U}^{(n)}  \approx \boldsymbol{\mathscr{X}}  
\end{equation}\normalsize

The advantages of \textit{Tucker} based factorization methods are already studied in several domains such as computer vision, \cite{vasilescu2002multilinear}, data mining \cite{savas2007handwritten}, and signal processing \cite{cong2015tensor,cichocki2015tensor}. However, in this research, we factorize tensor to obtain weights of convolution-tensorial filters for TFNet by devising our custom tensor factorization scheme which we call as Left one Mode Out Orthogonal Iteration (\textit{LoMOI}) presented in Alg.~\ref{Algo:HOSVD}. 

\begin{figure*}[t]
    \centering
    \captionsetup{justification=centerlast, skip=5pt}
        \subfloat[Convolution responses from the PCANet where each convolution response is visually distinct from the rest of the responses. This illustrates extraction of unique information with the amalgamated view of the data.]{\includegraphics[width=0.47\textwidth]{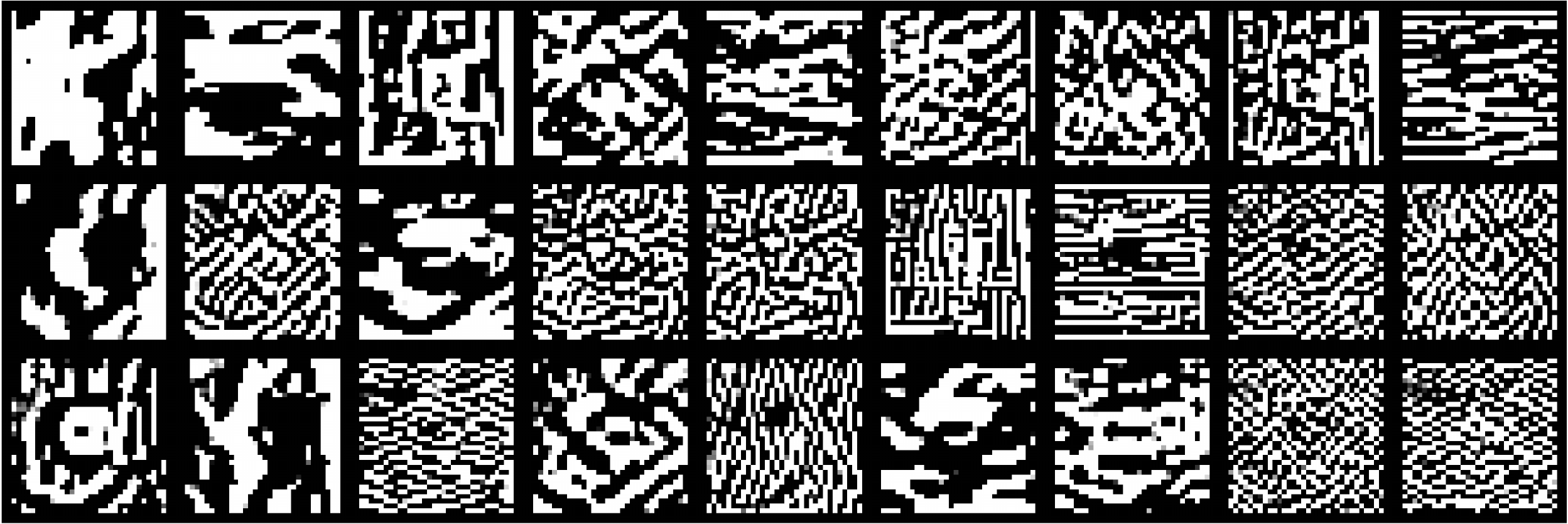}}
        \quad
        \subfloat[Convolution responses from the MPCANet. 
        Although the output is less diversified than TFNet, visual resemblance is partially observable.]{\includegraphics[width=0.47\textwidth]{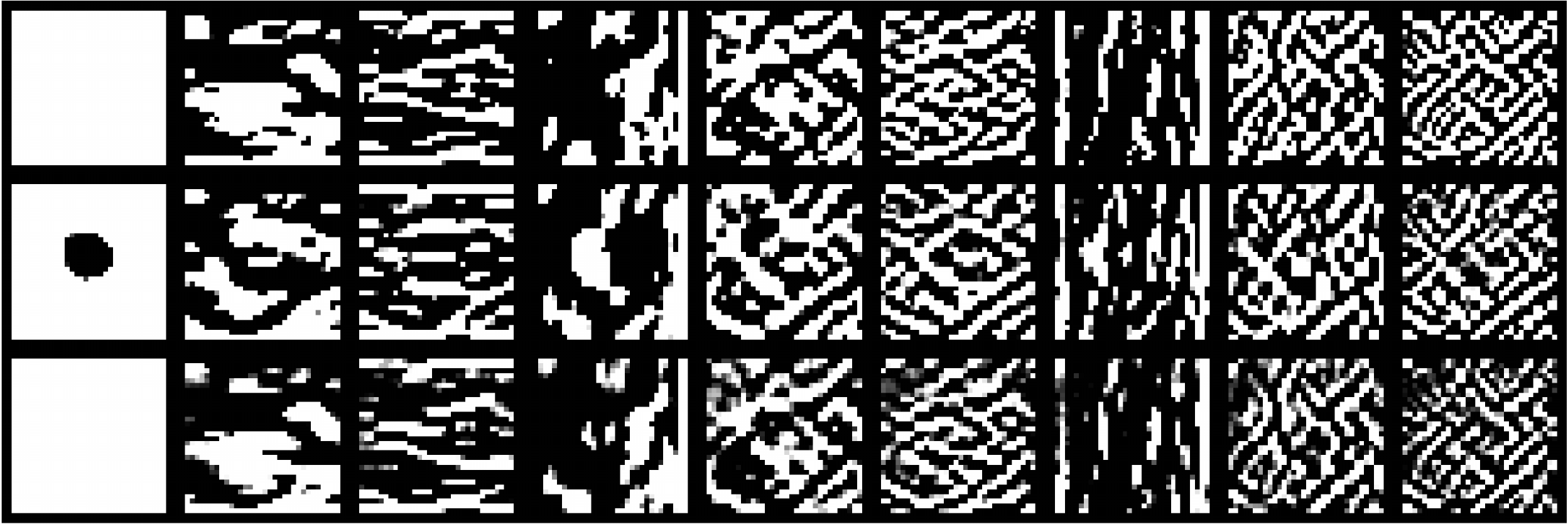}
        }
        \;
        \subfloat[Convolution responses from our TFNet where the visual resemblance is observed in a sequence of three responses. This illustrates extraction of common information with minutiae view of the data. ]{\includegraphics[width=0.47\textwidth]{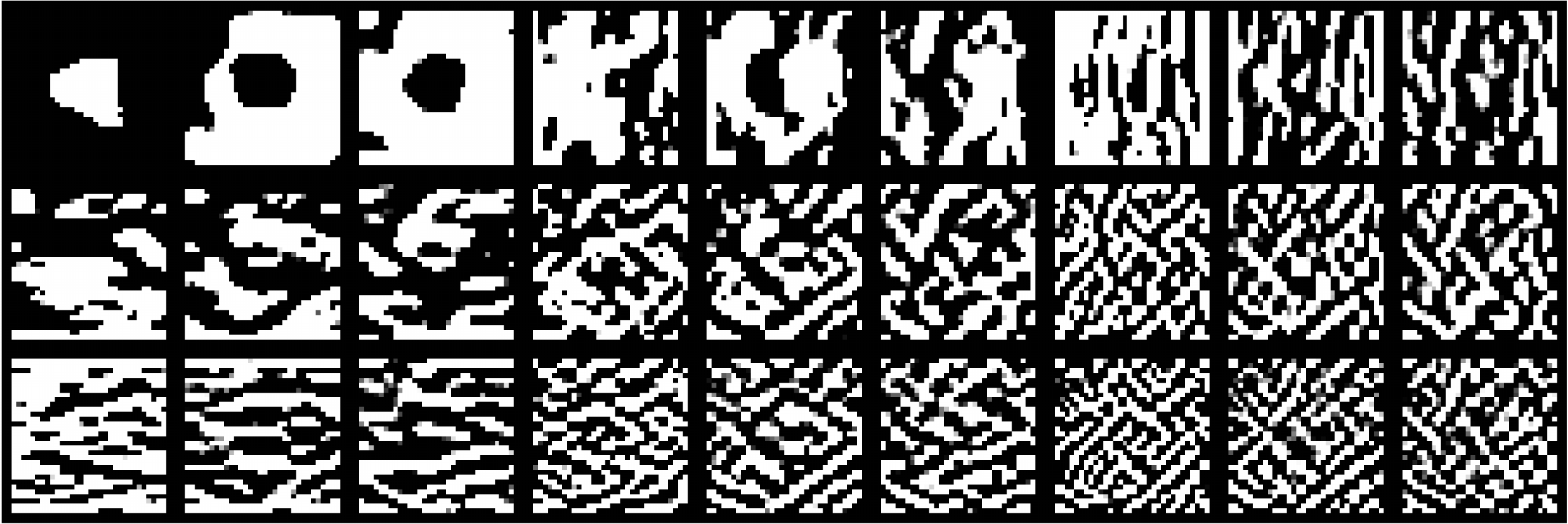}
        }
    \caption{Comparison of convolution outputs from Layer1 in PCANet, MPCANet and TFNet on CIFAR-10 dataset. These plots demonstrate the contrast between the kinds of information obtained with the amalgamated and the minutiae view of the data.}
\label{L1_Plot}
\vspace{-0.2cm}
\end{figure*}

\section{The Tensor Factorization Network}
\label{sec:TD}

The development of \textit{Tensor Factorization Network} (TFNet) is motivated to reduce the loss of spatial information occurring in the PCANet while vectorizing image patches. However, this transformation of the data is inherent while extracting the \emph{principal components} which destroys the geometric structure of the object encapsulated in the data which is proven beneficial in many image classification tasks \cite{vasilescu2002multilinear,verma2017extracting,chien2017tensor}. Furthermore, the vectorization of the data results in high dimensional vectors and generally requires more computational resources. Motivated by the above shortcomings with the PCANet, we propose the TFNet. The TFNet preserves the spatial structure of the data while obtaining weights of its convolution-tensor filters. The unsupervised feature extraction procedure with minutiae view of the data is detailed in the next subsection.

\subsection{The First Layer}
Similar to the first layer in PCANet, we begin by collecting all overlapping patches of size $\textit{k}_1 \times \textit{k}_2$ around each pixel from the image $\boldsymbol{I}_i$. However, contrary to PCANet the spatial structure of these patches are preserved and instead of matrix and we obtain a $3$-mode tensor $\boldsymbol{\mathscr{X}}_i \in \mathbb{R}^{\textit{k}_1 \times \textit{k}_2 \times \tilde{m}\tilde{n}}$. The mode-$1$ and mode-$2$ of this tensor represent the row-space, and the column-space spanned by the pixels in the image. Whereas the mode-$3$ of this tensor represents the total number of image patches obtained from the input image. Iterating this process for all the training images, we obtain $\boldsymbol{\mathscr{X}} \in \mathbb{R}^{k_1 \times k_2 \times N\tilde{m}\tilde{n}}$ as our final patch-tensor. The matrix factors utilized to generate our convolution-tensorial filters for to the first two modes of $\boldsymbol{\mathscr{X}}$ are obtained by utilizing our custom-designed \textit{LoMOI} (presented in Alg.~\ref{Algo:HOSVD}) in Eq.~\ref{tenfilt}.

\begin{equation} \label{tenfilt}
\footnotesize
[\hat{\boldsymbol{\mathscr{X}}}, \boldsymbol{U}^{(1)},\boldsymbol{U}^{(2)}] \gets LoMOI(\boldsymbol{\mathscr{X}}, r_1, r_2)
\end{equation}\normalsize
where $\hat{\boldsymbol{\mathscr{X}}} \in \mathbb{R}^{r_1 \times r_2 \times N\tilde{m}\tilde{n}}$, $\boldsymbol{U}^{(1)} \in \mathbb{R}^{\textit{k}_1 \times r_1}$, and $\boldsymbol{U}^{(2)} \in \mathbb{R}^{\textit{k}_2 \times r_2}$. We discard obtaining the matrix factors from mode-$3$ of tensor $\boldsymbol{\mathscr{X}}$ (which is $\boldsymbol{X}_3$) as this is equivalent to the transpose of the patches matrix  $\boldsymbol{X}$ in layer 1 of the PCANet which is not factorized in the PCANet while obtaining weights for its convolution filters. Moreover, the matrix factors for this mode span the sample space of the data which is trivial. A total of $L_1 = r_1 \times r_2$ convolution-tensor filters are obtained from the factor matrices $\boldsymbol{U}^{(1)}$ and $\boldsymbol{U}^{(2)}$ as in Eq.~\ref{eq:tenfilt}.
\begin{equation} \label{eq:tenfilt}
\footnotesize
 \boldsymbol{W}_{l_{TFNet}}^{1} = \boldsymbol{U}^{(1)}_{(:,i)} \otimes \boldsymbol{U}^{(2)}_{(:,j)} \in \mathbb{R}^{\textit{k}_1 \times \textit{k}_2}
\end{equation}
where `$\otimes$' is the \textit{outer}-product between two vectors, $i=[1,r_1]$, $j=[1,r_2]$, $l=[1,L_1]$, and $\boldsymbol{U}^{(\textit{m})}_{(:,i)}$ represents `$i^{th}$' column of the `$m^{th}$' factor matrix. Importantly, our convolution-tensorial filters do not require any explicit reshaping as the \textit{outer}-product between two vectors naturally results in a matrix. Therefore, we can straightforwardly convolve the input images with our obtained convolution-tensorial filters as described in Eq.~\ref{eq:tenconv} where $i=[1,N]$ and $l=[1,L_1]$ are the image and filter indices respectively.
\begin{equation} \label{eq:tenconv}
\footnotesize
 \boldsymbol{I}_{i_{TFNet}}^{l}  = \boldsymbol{I}_{i} \ast \boldsymbol{W}_{l_{TFNet}}^{1}
\end{equation}

However, whenever the data is an $RGB$-image, each extracted patch from the image is a $3$-order tensor $ \boldsymbol{\mathscr{X}} \in \mathbb{R}^{\textit{k}_1 \times \textit{k}_2 \times 3}$ (i.e., \textit{RowPixels}$\times$\textit{ColPixels}$\times${Color}). After collecting patches from all the training images, we obtain a $4$-mode tensor as $\boldsymbol{\mathscr{X}} \in \mathbb{R}^{{k}_1 \times \textit{k}_2 \times 3\times N\tilde{m}\tilde{n} }$ which is decomposed by utilizing \textit{LoMOI} ($[\hat{\boldsymbol{\mathscr{X}}}, \boldsymbol{U}^{(1)},\boldsymbol{U}^{(2)},\boldsymbol{U}^{(3)}] \gets LoMOI(\boldsymbol{\mathscr{X}}, r_1, r_2, r_3)$) for obtaining the convolution-tensorial filters in Eq.~\ref{eq:td13d}.
\begin{equation} \label{eq:td13d}
\footnotesize
\begin{split}
    & \qquad \boldsymbol{W}_{l_{TFNet}}^{1} = \boldsymbol{U}^{(1)}_{(:,i)} \otimes \boldsymbol{U}^{(2)}_{(:,j)} \otimes \boldsymbol{U}^{(3)}_{(:,k)}
\end{split}
\end{equation}
where $i \in [1,r_1]$, $j \in [1,r_2]$, and $k \in [1,r_3]$.

\subsection{The Second Layer}
Similar to the first layer, we extract overlapping patches from the input images and zero-center them to build a $3$-mode patch-tensor denoted as $\boldsymbol{\mathscr{Y}} \in \mathbb{R}^{k_1 \times k_2 \times N L_1\tilde{m}\tilde{n}}$ which is decomposed as $[\hat{\boldsymbol{\mathscr{Y}}}, \boldsymbol{V}^{(1)},\boldsymbol{V}^{(2)}] \gets LoMOI(\boldsymbol{\mathscr{Y}}, r_1, r_2)$ to obtain the convolution-tensor filters for layer 2 in Eq.~\ref{eq:td1}.
\begin{equation} \label{eq:td1}
\footnotesize
\begin{split}
    & \boldsymbol{W}_{l_{TFNet}}^{2} = \boldsymbol{V}^{(1)}_{(:,i)} \otimes \boldsymbol{V}^{(2)}_{(:,j)} \in \mathbb{R}^{\textit{k}_1 \times \textit{k}_2}
\end{split}
\end{equation}
where, $\hat{\boldsymbol{\mathscr{Y}}} \in \mathbb{R}^{r_1 \times r_2 \times N L_1\tilde{m}\tilde{n}}$, $\boldsymbol{V}^{(1)} \in \mathbb{R}^{\textit{k}_1 \times r_1}$, and $\boldsymbol{V}^{(2)} \in \mathbb{R}^{\textit{k}_2 \times r_2}$, $i=[1,r_1]$, $j=[1,r_2]$, and $l=[1,L_2]$. We, now convolve each of the $L_1$ input images from the first layer with the convolution-tensorial filters obtained as below in Eq.~\ref{eq:tenconv2}. 
\begin{equation} \label{eq:tenconv2}
\footnotesize
 \boldsymbol{O}_{i_{TFNet}}^{l}  = \boldsymbol{I}_{i_{TFNet}}^l \ast \boldsymbol{W}_{l_{TFNet}}^{2},  \quad	  l=1,2,...,L_2
\end{equation}

The number of output images obtained here is equal to $L_1 \times L_2$ which is identical to the number of images obtained at layer 2 of PCANet. Finally, we utilize the output layer of PCANet (Sec.~\ref{ol}) to obtain the feature vectors from the minutiae view of the image in Eq.~\ref{eq:hashtd}.
\begin{equation} \label{eq:hashtd}
\footnotesize
\begin{split}
    & \qquad \qquad \mathcal{\boldsymbol{I}}_{i_{TFNet}}^{l}  = \sum_{l=1}^{L_2} 2^{l-1}H(\boldsymbol{O}_{l_{TFNet}}^2) \\
    &  f_{i_{TFNet}}  =[Bhist( \mathcal{\boldsymbol{I}}_{i_{TFNet}}^{1}), ..., Bhist( \mathcal{\boldsymbol{I}}_{i_{TFNet}}^{L_1})]^T \in \mathbb{R}^{(2^{L_2})L_1B} 
\end{split}
\end{equation}

\begin{figure*}[t]
\begin{center}
\captionsetup{justification=centering}
\includegraphics[width = 0.9\textwidth]{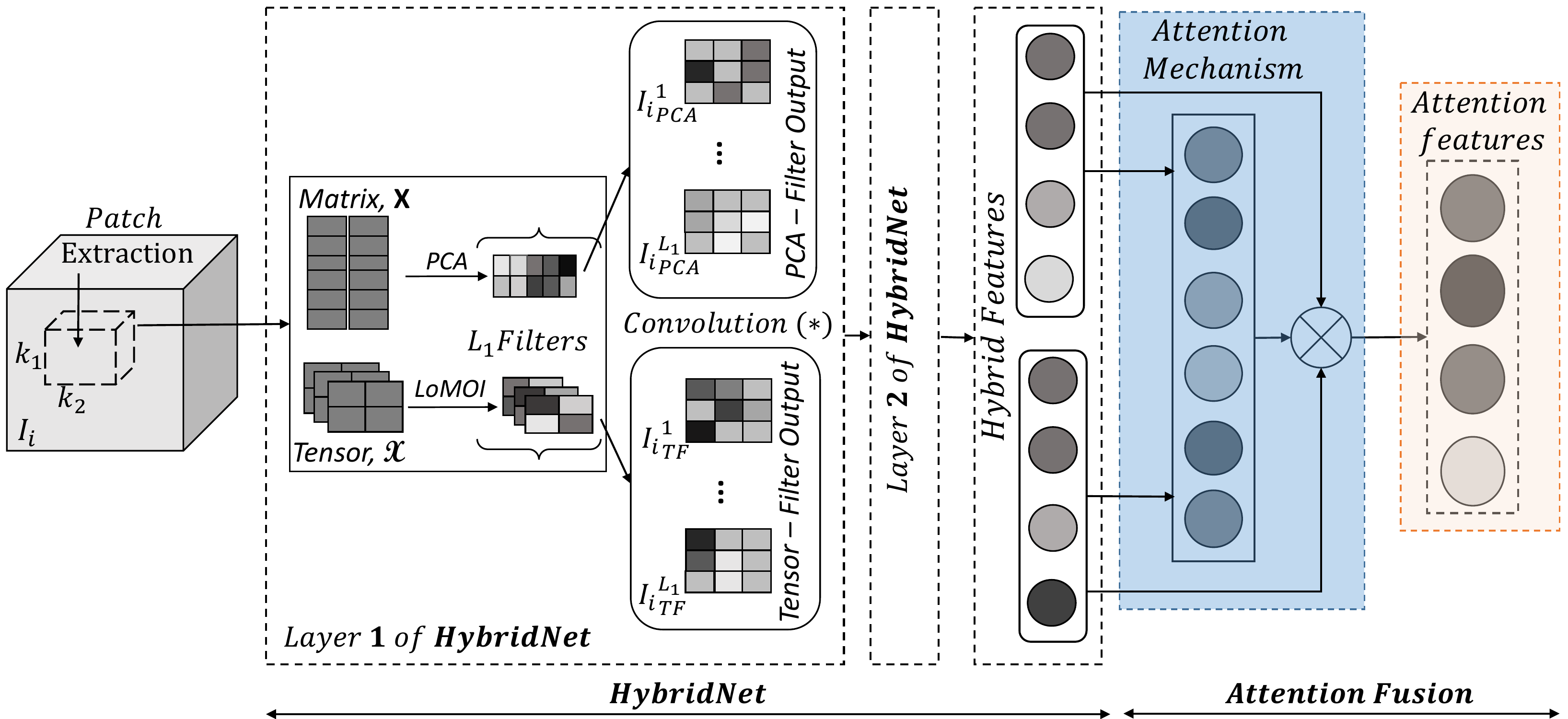}
\setlength{\belowcaptionskip}{-0.3cm}
\caption{Workflow of the proposed \textit{Attn-HybridNet} model.}
\label{fig:hd}
\end{center}
\end{figure*}

Despite having close resemblance between the feature extraction mechanism of the PCANet and the TFNet, these two networks capture visibly distinguishable features from the two view of the images as shown in Fig.~\ref{L1_Plot}. These plots are obtained by convolving image of a $cat$ with the convolution filters obtained in the first layer of the networks. 

Undoubtedly, each of the $L_1$ convolution responses within the PCANet is visibly distinct. Whereas the convolution responses within the TFNet shows visual similarity, i.e., the images in a triplet sequence show similarity consecutively. These plots demonstrate that the TFNet emphasize mining the \textit{common} information from the minutiae view of the data. Whereas the PCANet emphasizes mining the \textit{unique} information from the amalgamated view of the data. Both these kinds of information are proven beneficial for classification in \cite{liu2013mining,verma2017extracting} and motivate the development of \textit{HybridNet}.

Besides, we also present convolution responses of MPCANet \cite{wu2017multilinear} in the comparisons. MPCANet employs tensorized-convolution but differs from our TFNet in two ways: 1) construction of convolution kernels and 2) convolution operation. Technically, the convolution kernel and convolution operation in MPCANet in \cite{wu2017multilinear} (and its predecessor in \cite{zeng2015tensor}) are combined together as conventional tensor operations, obtaining a) factor matrices for each mode from the patch-tensor and b) n-mode products of factor matrices with the patch-tensor. In other words, the convolution in MPCANet is a n-mode product of the patch-tensor and factor matrices, whereas in TFNet the convolution kernels are obtained by performing outer-procut of factor matrices in Eq.~\ref{eq:tenfilt} followed by convolution in Eq.~\ref{eq:tenconv}. From \Cref{L1_Plot}, it is  visible that the convolution responses in MPCANet is less diversified than those of TFNet although visible resemblance is still observable. We believe that this is because of how the convolution filters and convolution operation are performed in MPCANet as analyzed above. 

\section{ The Hybrid Network}
\label{sec:HD}

The PCANet and the TFNet extract contrasting information from the amalgamated view and the minutiae view of the data, respectively. However, we hypothesize that the information from both these views are essential as they conceal complementary information and that their integration can enhance the performance of classification systems. Motivated by the above, we propose the \textit{HybridNet}, which simultaneously extracts information from both views of the data and is detailed in the next subsection. However, for ease of understanding, we illustrate the complete procedure of feature extraction with \textit{Attn-HybridNet} in Fig.~\ref{fig:hd}.

\subsection{The First Layer}
Similar to the previous networks, we begin the feature extraction process by collecting all overlapping patches of size $\textit{k}_1 \times \textit{k}_2$ around each pixel from the image $I_i$. Importantly, the first layer of \textit{HybridNet} consists of image-patches expressed both as tensors $\boldsymbol{\mathscr{X}} \in \mathbb{R}^{{k}_1 \times \textit{k}_2 \times 3\times N\tilde{m}\tilde{n} }$ and matrices $\boldsymbol{X} \in \mathbb{R} ^{ k_1 k_2 \times N\tilde{m}\tilde{n}} $ which are utilized for obtaining weights of convolution filters in layer 1 of \textit{HybridNet}. 

\begin{figure}[t]
\captionsetup{justification=centering}
    \centering
        \subfloat[Comparison of eigenvalues between networks]{\includegraphics[width=0.42\columnwidth]{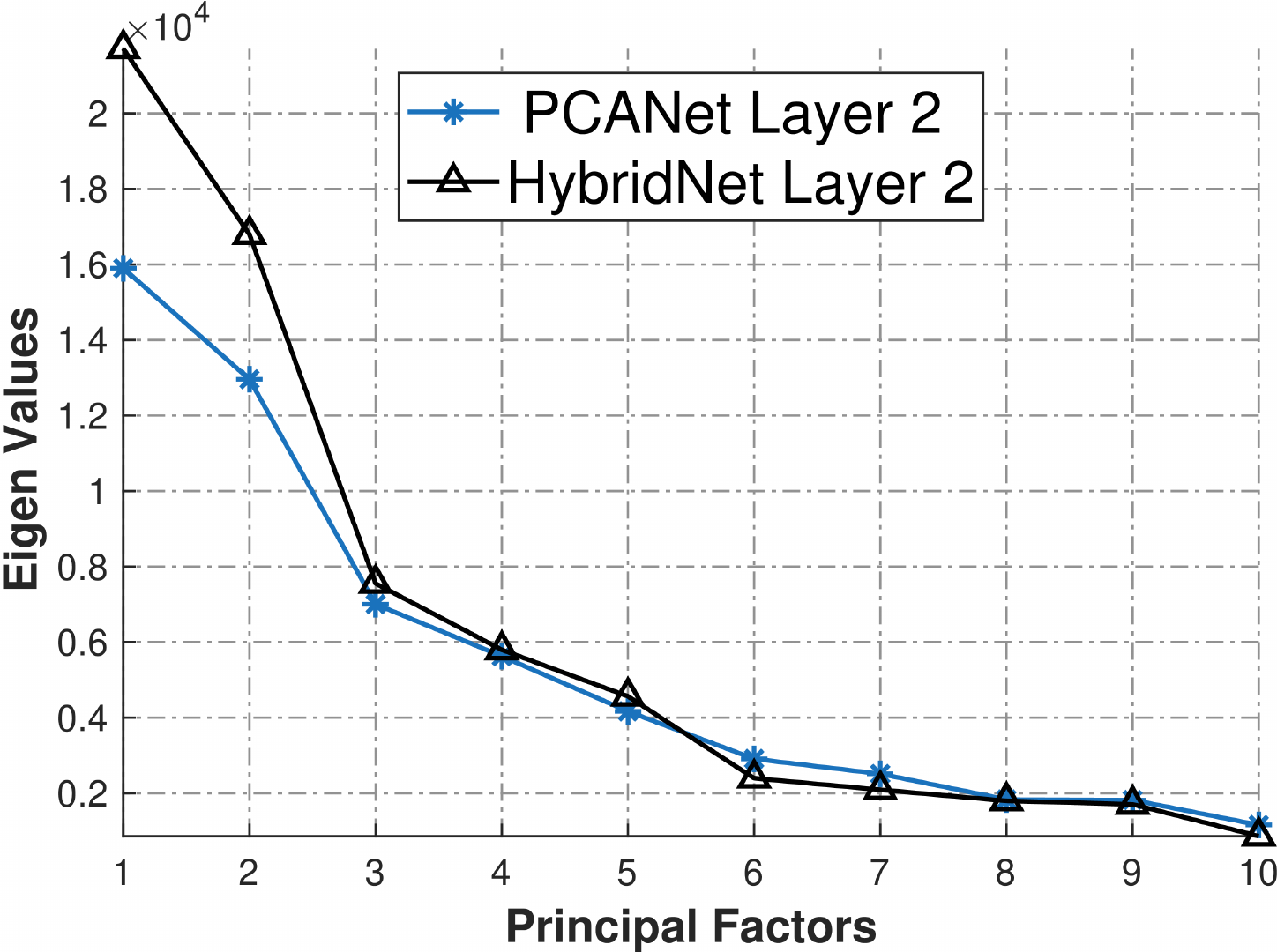}}
        \quad
        \subfloat[Comparison of core-tensor strength between networks]{\includegraphics[width=0.48\columnwidth]{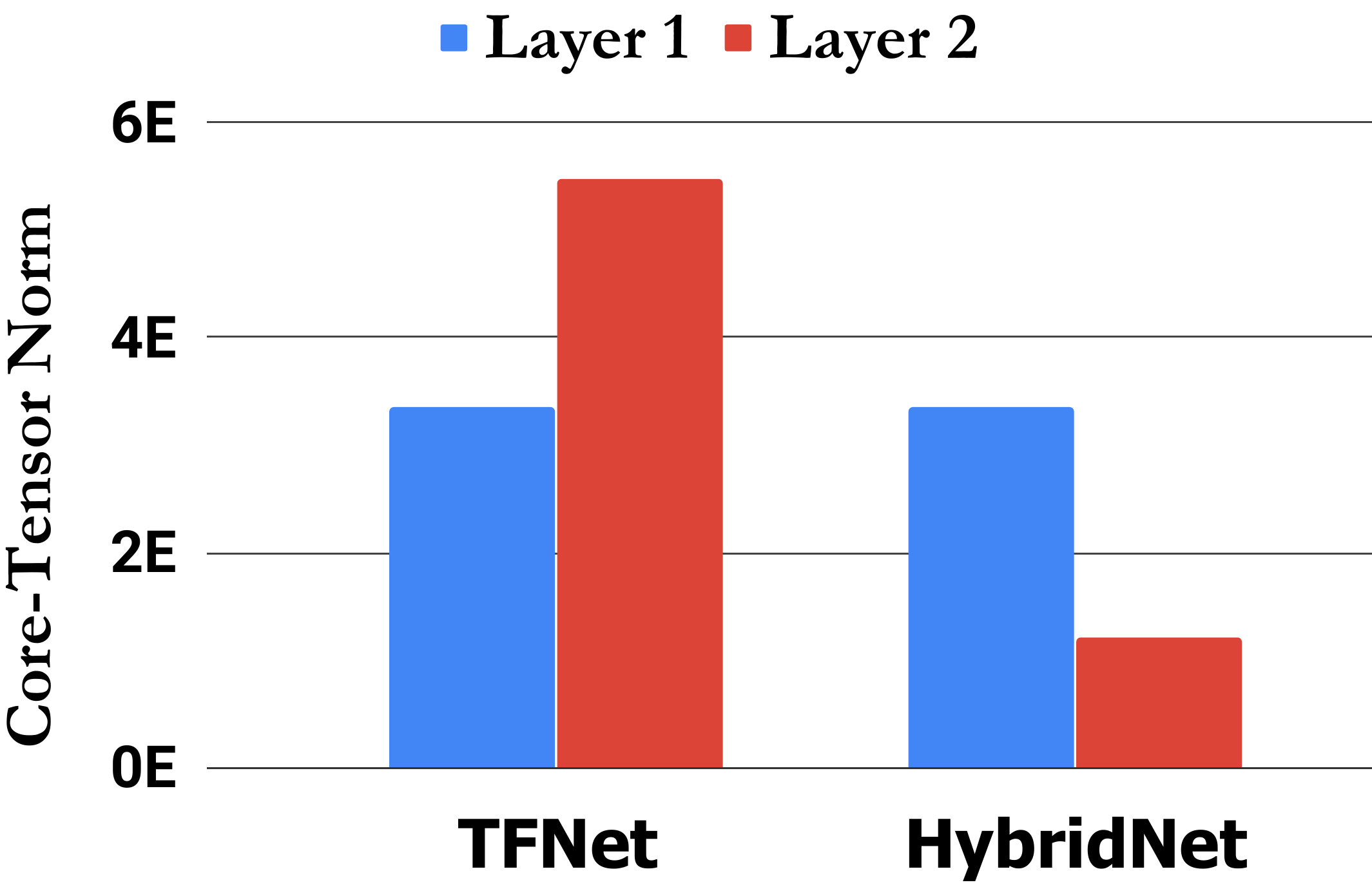}}
    \setlength{\belowcaptionskip}{-0.2cm}
    \caption{Comparison of factorization strength in Layer 2 of the PCANet, TFNet and \textit{HybridNet} on CIFAR-10 dataset}
\label{comp_eig}
\end{figure}

This enables this layer (and the subsequent layers) of \textit{HybridNet} to learn superior filters as they perceive more information from both views of the data. The weights for the \textit{pca}-filters are obtained as the principal-eigenvectors as $\boldsymbol{W}_{l_{PCA}}^{1} = mat_{k_1,k_2} (ql(\boldsymbol{X} \boldsymbol{X}^{T}))$, and the weights for convolution-tensor filters are obtained by utilizing \textit{LoMOI} as $ \boldsymbol{W}_{l_{TF}}^{1} = \boldsymbol{U}^{(1)}_{(:,i)} \otimes \boldsymbol{U}^{(2)}_{(:,j)} \otimes \boldsymbol{U}^{(3)}_{(:,k)} $. Furthermore, the output from this layer is obtained by convolving input images with a) the PCA-filters and b) the convolution-tensorial filters in Eq.~\ref{eq:hd_pcafilt}. This injects more diversity to the output in succeeding layer of \textit{HybridNet}.
\begin{equation} \label{eq:hd_pcafilt}
\footnotesize
\begin{split}
    & \qquad  \boldsymbol{I}_{i_{PCA}}^{l}  = \boldsymbol{I}_{i} \ast \boldsymbol{W}_{l_{PCA}}^{1} \\
    & \qquad \boldsymbol{I}_{i_{TF}}^{l}  = \boldsymbol{I}_{i} \ast \boldsymbol{W}_{l_{TF}}^{1}
\end{split}
\end{equation}
Since we obtain of $L_1$ \textit{pca} filters and $L_1$ convolution-tensor filters, a total of $2 \times L_1$ outputs are obtained in this layer.

\subsection{The Second Layer}

Similar to the first layer, we begin with collecting all overlapping patches of size $\textit{k}_1 \times \textit{k}_2$ around each pixel from the images. However, contrary to the above layer, the weights of the \textit{pca}-filters $\boldsymbol{W}_{l_{PCA}}^{2}$ and convolution-tensor filters $\boldsymbol{W}_{l_{TF}}^{2}$ are learned from the data obtained by convolving input images with the \textit{pca} filters and the convolution-tensor filters i.e. both $\boldsymbol{I}_{i_{PCA}}$ and $\boldsymbol{I}_{i_{TF}}$. Hence both the patch-matrix $\boldsymbol{Y} \in  \mathbb{R} ^{ k_1 k_2 \times 2 L_1 N\tilde{m}\tilde{n}}$ and the patch-tensor $ \boldsymbol{\mathscr{Y}} \in \mathbb{R}^{k_1 \times k_2 \times 2 L_1 N \tilde{m}\tilde{n}} $ contain image patches obtained from $[\boldsymbol{I}_{i_{PCA}} , \boldsymbol{I}_{i_{TF}}]$. This enables the hybrid filters to assimilate more variability present in the data while obtaining weights of their convolution filters as evident in Fig.~\ref{comp_eig}.

The plot in Fig.~\ref{comp_eig}(a) compares the eigenvalues obtained in layer 2 (we exclude eigenvalues from layer 1 as they completely overlap as their expected behavior). The leading eigenvalues obtained in layer 2 of the \textit{HybridNet} by $principal$ $components$ has much higher magnitude than the corresponding eigenvalues obtained by $principal$ $components$ in PCANet. This demonstrates that the $pca$ filters in the \textit{HybridNet} capture more variability than those in the PCANet.

\begin{algorithm}[t]    
\scriptsize
\captionsetup{justification=justified, skip=5pt}
\caption{The \textit{HybridNet} Algorithm}
\begin{algorithmic}[1]
\State\textbf {Input:} $\boldsymbol{I}_{i}, i=1,2,...,n$ n is the total number of training images, $L = [l_{1}, l_{2},...l_{D}]$ the number of filters in each layer,$k_{1}$ and $k_{2}$ the patch-size, $B$, $D = $ the depth of the network.

\For {$i =1,2,...,n$ }     \Comment{DO for each image in the first convolution layer.}
\State $ \boldsymbol{X} \gets $ extract patches of size $k_1 \times k_2$ around each pixel of $I_i$ \Comment{mean centered and vectorized.}
\State $ \boldsymbol{\mathscr{X}} \gets $ extract patches of size $k_1 \times k_2$ around each pixel of $I_i$ \Comment{mean centred but retain their spatial shape.}
\EndFor 

\State $\boldsymbol{W}_{PCA} \gets $ obtain PCA filters by factorizing $\boldsymbol{X}$.
\State $\boldsymbol{W}_{TF} \gets$ obtain tensor filters by factorizing $\boldsymbol{X}$ with LoMOI Algo.~\ref{Algo:HOSVD}.

\For {$i =1,2,...,n$}    
\State $ \boldsymbol{I}_{i_{PCA}}^{1} \gets \boldsymbol{I}_{i} \ast \boldsymbol{W}_{PCA}$ \Comment{store convolution with pca filters.}
\State $ \boldsymbol{I}_{i_{TF}}^{1}  \gets \boldsymbol{I}_{i} \ast \boldsymbol{W}_{TF} $ \Comment{store convolution with tensorial filters.}
\EndFor 

\For{$l = 2,...,D$} \Comment{DO for the remaining convolution layers.}

\State $ \boldsymbol{I} \gets [\boldsymbol{I}_{{PCA}}^{(l-1)},  \boldsymbol{I}_{{TF}}^{(l-1)}] $ \Comment{Utilize both views together.}
\For {$i =1,2,...,\hat{n}$ }     \Comment{$\hat{n} = 2 \times n\times l_{l-1} $.}
\State $ \boldsymbol{X} \gets $ extract patches of size $k_1 \times k_2$ around each pixel of $\boldsymbol{I}_i$ 
\State $ \boldsymbol{\mathscr{X}} \gets $ extract patches of size $k_1 \times k_2$ around each pixel of $\boldsymbol{I}_i$ 
\EndFor 
\State $\boldsymbol{W}_{PCA}^{l} \gets $ obtain PCA filters by factorizing $\boldsymbol{X}$.
\State $\boldsymbol{W}_{TF}^{l} \gets$ obtain tensor filters by factorizing $\boldsymbol{X}$ with LoMOI Alg.~\ref{Algo:HOSVD}.

\For {$i =1,2,...,\bar{n}$}    \Comment{$\bar{n} = n\times l_{l-1} $.}
\State $ \boldsymbol{I}_{i_{PCA}}^{l}  =  \boldsymbol{I}_{i_{PCA}}^{(l-1)} \ast \boldsymbol{W}_{l_{PCA}}^{1}$ 
\State $ \boldsymbol{I}_{i_{TF}}^{l}    = \boldsymbol{I}_{i_{TF}}^{l-1} \ast \boldsymbol{W}_{l_{TF}}^{1} $ 
\EndFor 

\EndFor

\For {$i =1,2,...,n$ }     \Comment{DO for each image.}
\State $\mathcal{\boldsymbol{I}}_{i_{PCA}}  = \sum_{k=1}^{l_d} 2^{l-1}H(\boldsymbol{I}_{k_{PCA}}^{d}) $ \Comment{Binarize and accumulate output from all convolution layers.}
\State $\mathcal{\boldsymbol{I}}_{i_{TF}}  = \sum_{k=1}^{l_d} 2^{l-1}H(\boldsymbol{I}_{k_{TF}}^{d}) $ 
\State $ f_{i_{PCA}} \gets Bhist(\mathcal{\boldsymbol{I}}_{i_{TF}}) $   \Comment{create block-wise histogram.}
\State $  f_{i_{TF}} \gets Bhist(\mathcal{\boldsymbol{I}}_{i_{TF}}) $
\EndFor

\State\textbf {Output:} features from the amalgamated mode as $f_{PCA}$, and the minutiae mode as $f_{TF}$.  

\end{algorithmic}

\label{Algo:hd}
\end{algorithm}

Similarly, Fig~\ref{comp_eig}(b) compares the core-tensor strength in different layers of the \textit{HybridNet} and the TFNet. We plot the norm of the core-tensor for both the networks as the values in the core-tensor is analogous to eigenvalues for higher-order matrices, and its norm signifies the compression strength of the factorization \cite{lu2013multilinear}. Again, the norm of the core-tensor in layer 2 of \textit{HybridNet} is much lower than that of the TFNet, suggesting relatively higher factorization strength in \textit{HybridNet}. Besides, as expected, the norm of the core-tensor in layer 1 for both the networks coincides and signifies equal factorization strength at this layer. Consequently, this leads to attainment of better-disentangled feature representations with the \textit{HybridNet} and hence enhances its generalization performance over the PCANet and the TFNet by integrating information from the two views of the data. 


In the second layer, the weights of \textit{pca} filters are obtained by \textit{principal components} as $\boldsymbol{W}_{l_{PCA}}^{2} = mat_{k_1,k_2} (ql(\boldsymbol{Y} \boldsymbol{Y}^{T}))$ and the weights for convolution-tensor filters are obtained as $ \boldsymbol{W}_{l_{TF}}^{2} = \boldsymbol{V}^{(1)}_{(:,i)} \otimes \boldsymbol{V}^{(2)}_{(:,j)}$, where the matrix factors are obtained using \textit{LoMOI} $[\hat{\boldsymbol{\mathscr{Y}}}, \boldsymbol{V}^{(1)},\boldsymbol{V}^{(2)}] \gets LoMOI(\boldsymbol{\mathscr{Y}}, r_1, r_2)$. Analogous to the previous networks, the output images from this layer of \textit{HybridNet} are obtained by a) convolving the $L_1$ images corresponding to the output from the PCA-filters in the first layer with the $L_2$ $pca$ filters obtained in the second layer (Eq.~\ref{eq:hd_pca2}), and b) convolving the $L_1$ images corresponding to the output from the convolution-tensorial filters in the first layer with the $L_2$ convolution-tensorial filters obtained in the second layer (Eq.~\ref{eq:hd_td2}). This generates a total of $2 \times L_1 \times L_2$ output images in this layer. 
\begin{equation} \label{eq:hd_pca2}
\footnotesize
\begin{split}
    & \boldsymbol{O}_{i_{PCA}}^{l}  = \boldsymbol{I}_{i_{PCA}}^l \ast \boldsymbol{W}_{l_{PCA}}^{2}
\end{split}
\end{equation}
\begin{equation} \label{eq:hd_td2}
\footnotesize
\begin{split}
    & \boldsymbol{O}_{i_{TF}}^{l}  = \boldsymbol{I}_{i_{TF}}^l \ast \boldsymbol{W}_{l_{TF}}^{2}
\end{split}
\end{equation}
The output images obtained from the \textit{pca}-filters ($\boldsymbol{O}_{i_{PCA}}^{l}$) in layer 2 are then processed with the output layer of the PCANet (Sec.~\ref{ol}) to obtain $f_{i_{PCA}}$ as the information from amalgamated view of the image. Similarly, the output images obtained from the convolution-tensor filters ($\boldsymbol{O}_{i_{TF}}^{l}$) are processed to obtain $f_{i_{TF}}$ as the information from minutiae view of the image. Finally, these two kinds information are concatenated to obtain the hybrid features as in Eq.~\ref{hdfeat}.
\begin{equation} \label{hdfeat}
\footnotesize
\begin{split}
    & f_{i_{hybrid}}  = [ f_{i_{PCA}} \ f_{i_{TF}}] \in \mathbb{R}^{(2^{L_2})2L_1B}
\end{split}
\end{equation}

The whole procedure of obtaining hybrid features is detailed in Alg.~\ref{Algo:hd}. Although these hybrid features bring the best of both the common and the unique information obtained respectively from the two views of the data, they still suffer from feature redundancy problems induced by the spatial pooling operation in the output layer. To alleviate this drawback, we propose the \textit{Attn-HybridNet}, which further enhances the discriminability of the hybrid features.   


\section{attention-based fusion - {Attn-HybridNet}} \label{prop_attn}

Our proposed \textit{HybridNet} eradicates the loss of information by integrating the learning scheme of PCANet and TFNet thus obtaining superior features than either of the networks. However, the feature encoding scheme in the output layer is elementary and induces redundancy in the feature representations \cite{jia2013compact,coates2011importance}. Moreover, the generalized spatial pooling operation in the output layer is unable to accommodate the spatial structure of the natural images, i.e., it is more effective for aligned images dataset like face and handwritten digits than for object recognition dataset. Simply, the design of the output layer is ineffectual to obtain utmost feature representation on object recognition datasets resulting in performance degradation with the \textit{HybridNet}. Moreover, efficient ways to alleviate this drawback with the output layer are not addressed in the literature, which necessitates the development of our proposed attention-based fusion scheme i.e. the \textit{Attn-HybridNet}. 

\begin{algorithm}[t]    
\scriptsize
\captionsetup{justification=justified, skip=5pt}
\caption{The \textit{Attn-HybridNet} Algorithm}
\begin{algorithmic}[1]
\State\textbf {Input:} $f_{{hybrid}}  = [ f_{{PCA}}; \ f_{{TF}}] \in \mathbb{R}^{N \times (2^{L_2})L_1B \times 2}$ the hybrid feature vectors from the training images; $y = [0,1,...,C]$ ground truth of training images, dimensionality of feature level context vector $w \in \mathbb{R}^d$, where $d << \mathbb{R}^{(2^{L_2})L_1B}$.

\State randomly initialize $\boldsymbol{W}$, $ f_{c}$, and $w$  
\State $loss \gets  1000$ \Comment{arbitrary number to start training}

\Do
\State $[f_{batch}, y_{batch}] \gets $ sample batch $([f_{{\textbf{hybrid}}},y])$
\State $\boldsymbol{P}_{F} \gets tanh(W.f_{batch})$ \Comment{get the hidden representation of the hybrid features}
\State $\alpha = softmax \big( w^{T}.\boldsymbol{P}_{F} \big)$ \Comment{measure and normalize the importance}
\State $F_{attn} =  f_{batch}.\alpha^{T}$ \Comment{perform attention fusion}
\State $\hat{y} \gets f_{c}(F_{attn})$ \Comment{fully connected layer}
\State $loss \gets  LogLoss(y_{batch},\hat{y}_{batch})$ \Comment{compute loss for optimizing parameters}
\State back-propagate loss for optimizing $\boldsymbol{W}$, $ f_{c}$, and $w$.

\doWhile{$[(loss \ge \varepsilon)  ]$} \Comment{loop until convergence}

\State\textbf {Output:} parameters to perform attention fusion $\boldsymbol{W}$, $f_{c}$, and $w \in \mathbb{R}^d$  

\end{algorithmic}

\label{Algo:Attn}
\end{algorithm}

Our proposed attention-based fusion scheme is presented in Alg.~\ref{Algo:Attn}, where $f_{hybrid} \in \mathbb{R}^{N \times (2^{L_2})L_1B \times 2} $ are the hybrid feature vectors obtained with the \textit{HybridNet}, $w \in \mathbb{R}^d$ is the feature level context vector of dimension $d << (2^{L_2})L_1B$, $\alpha^{T} \in \mathbb{R}^2$ is the normalized importance weight vector for combining the two kinds of information with attention fusion, and $F_{attn} \in \mathbb{R}^{(2^{L_2})L_1B} $ are the attention features. The fully connected layers i.e. $\boldsymbol{W} \in \mathbb{R}^{d \times (2^{L_2})L_1B}$ and $f_{c}$ are utilized to obtain hidden representations of features while performing attention fusion.

A few numerical optimization based techniques proposed in \cite{jia2013compact,jia2012beyond} exist for alleviating the feature redundancy from architectures utilizing generalized spatial pooling layers. However, these techniques require grid search between the dictionary size (number of convolution filters in our case) and the pooling blocks in the output layer while performing optimization. Besides, the transition to prune filters from a single-layer networks to multi-layer network is not smooth in these techniques. A major difference between our proposed \textit{Attn-HybridNet} and the existing proposal in \cite{jia2013compact,jia2012beyond} is that we reduce the feature redundancy by performing feature selection with attention-based fusion scheme,  whereas the existing techniques prune the filters to eliminate the feature redundancy. Therefore, our proposed \textit{Attn-HybridNet} is superior to these existing techniques as it decouples the two sub-processes, i.e., information discovery with convolution layers and feature aggregation in the pooling layer while alleviating the redundancy exhibiting in the feature representations.

The discriminative features obtained by \textit{Attn-HybridNet} i.e. $F_{attn}$ are utilized with $softmax$-layer for classification, where the parameters in the proposed fusion scheme (i.e., $\boldsymbol{W}$, $f_{c}$ and $w$) are optimized via gradient-descent on the classification loss. This simple yet effective scheme substantially enhances the classification performance by obtaining highly discriminative features. Comprehensive experiments are conducted in this regard to demonstrate the superiority of \textit{Attn-HybridNet} detailed in Sec.~\ref{sec:exp}.

\subsection{Computational Complexity}

To calculate the computational complexity of \textit{Attn-HybridNet}, we assume the \textit{HybridNet} is composed of two-layers with a patch size of $k_1 = k_2 = k$ in each layer followed by our attention-based fusion scheme.

In each layer of the \textit{HybridNet}, we have to compute the time complexities arising from learning convolution weights from the two views of the data. The formation of the zero-centered patch-matrix $\boldsymbol{X}$ and zero-centered patch-tensor $\boldsymbol{\mathscr{X}}$ have identical complexities as $ k^2 (1 + \tilde{m} \tilde{n})$. The complexity of eigen-decomposition for patch-matrix and tensor factorization with \textit{LoMOI} for patch-tensor are also identical and equal to $\mathcal{O}((k^2)^3)$, where $k$ is a whole number $<$ 7 in our experiments. Further, the complexity for convolving images with the convolution filters at stage $i$ requires $L_i k^2mn$ flops. The conversion of $L_2$ binary bits to a decimal number in the output layer costs $2L_2\tilde{m}\tilde{n}$, where $\tilde{m} = m - \lceil{\frac{k}{2}}\rceil,  \tilde{n} = n - \lceil{\frac{k}{2}}\rceil$ and the naive histogram operation for this conversion results in complexity equal to $\mathcal{O}(mnBL_2log 2)$.

The complexity of performing matrix multiplication in \textit{Attn-HybridNet} is $\mathcal{O}\Big( 2L_1B \big( d( 1 + 2^{L_2} ) + 2^{L_2} \big) \Big)$ which can be efficiently handled with modern deep learning packages like Tensorflow \cite{abadi2016tensorflow} for stochastic updates. To optimize the parameters in the attention-based fusion scheme ($\boldsymbol{W}$, $f_c$, and $w$), we back-propagate the loss through the attention network until convergence of the error on the training features.


\section{Experiments and Results }
\label{sec:exp}

\subsection{Experimental Setup}

In our experiments, we utilized a two-layer architecture for each of the networks in comparison, while the number of convolution filters in the first and the second layer are optimized via cross-validation on the training datasets. The dimensionality of the feature vectors extracted from PCANet and TFNet is then $B L_1 2^{L_2}$, where $L_1$ and $L_2$ are the number of convolution filters in layer 1 and layer 2 respectively. The dimensionality of feature vector with \textit{HybridNet} is then $2 B L_1 2^{L_2}$. We utilized \emph{Linear-SVM} \cite{fan2008liblinear} as the classifier incorporating features obtained with the PCANet, TFNet, and the \textit{HybridNet}. 

The attention-based fusion scheme is performed by following the procedure as described in Alg.~\ref{Algo:Attn}, where we obtained the optimal attention dimension on training data for the context level feature vector $w \in \mathbb{R}^{d}$ in $[10,50,100,150,200,400]$. The obtained attention features i.e. $ F_{attn} \in R^{B L_1 2^{L_2}}$ are utilized with $softmax$-layer for classification. The parameters of attention-based fusion scheme ($\boldsymbol{W}$, $f_c$, and $w$) are optimized via back-propagation on the classification loss implemented in TensorFlow \cite{abadi2016tensorflow}. We observed that the attention-network's optimization took less than 15 epochs for convergence on all the datasets utilized in this paper. 

\begin{figure*}[t]
    \centering
    \captionsetup{justification=centerlast, skip=5pt}
        \subfloat[CIFAR-10]{\includegraphics[width=0.31\textwidth]{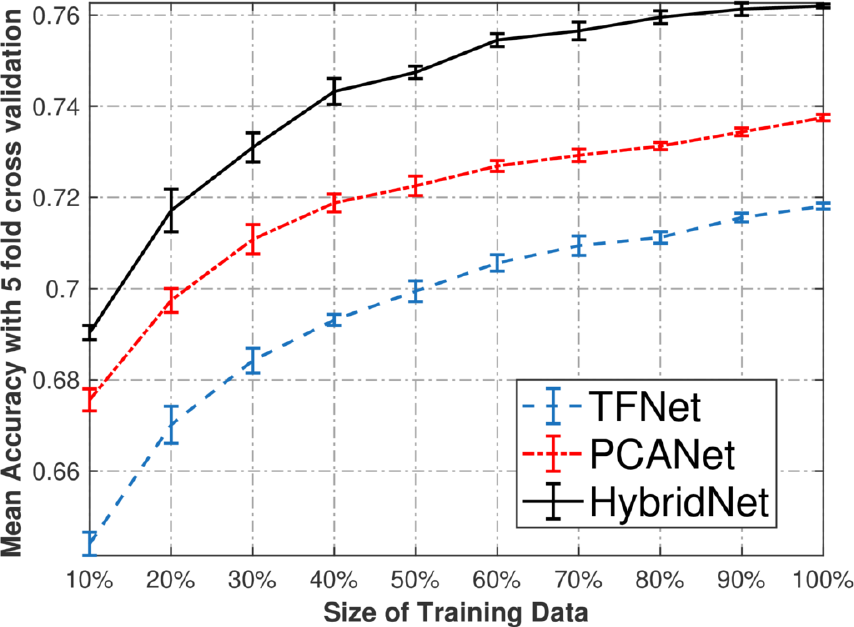}}
        \quad
        \subfloat[MNIST Rotation ]{\includegraphics[width=0.31\textwidth]{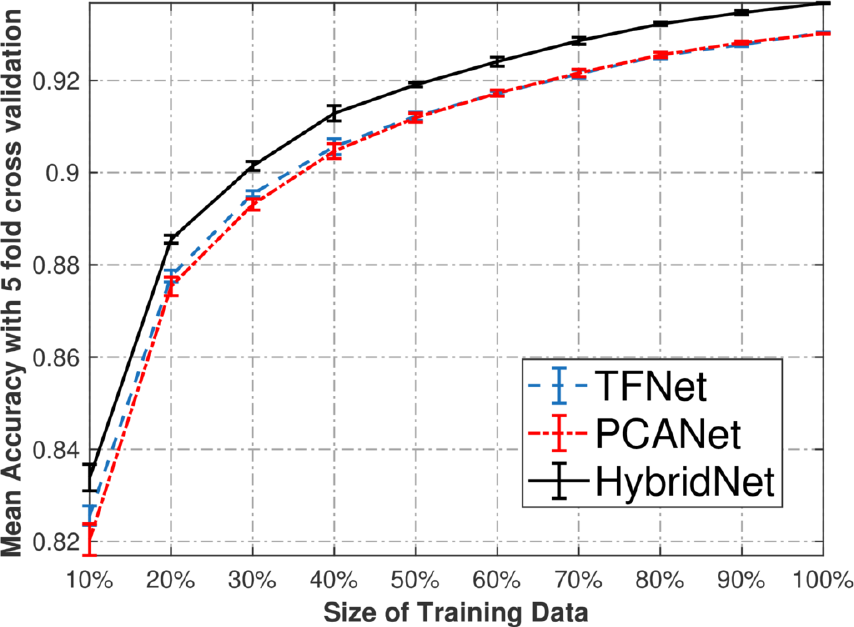}}
        \quad
        \subfloat[MNIST bg-img-rot ]{\includegraphics[width=0.31\textwidth]{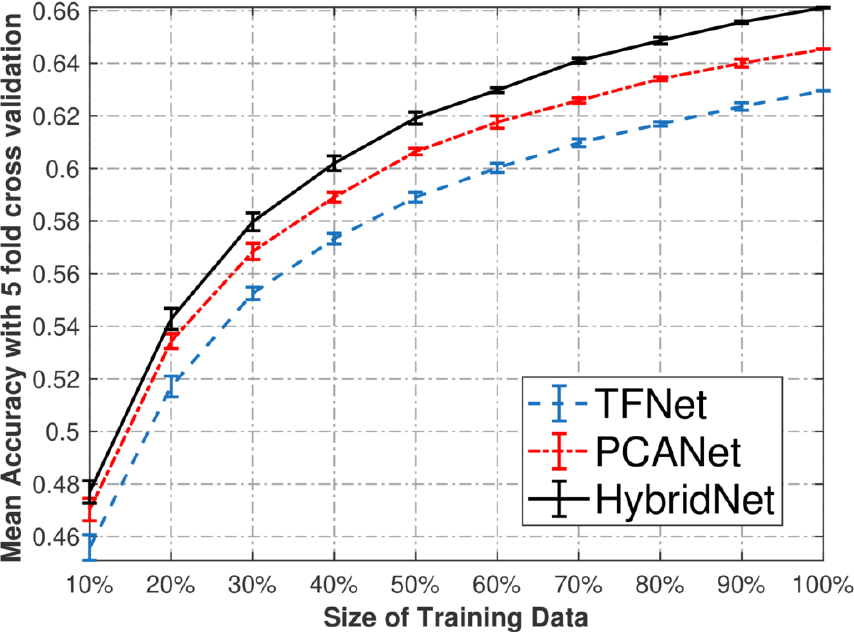}}
    \caption{Performance Comparison by varying size of the training data}
\label{Q1}
\end{figure*}

\subsection{Datasets}
\label{sec:base}

The details of datasets and hyper-parameters are as below:
\begin{itemize}

\item [\textbf{1}] MNIST variations \cite{larochelle2007empirical}, which consist of gray scale handwritten digits of size  $28 \times 28$ with controlled factors of variations such as background noise, rotations, etc. Each variation contains $10K$ training and $50K$ testing images. We cite the results for baselines techniques like 2-stage ScatNet \cite{bruna2013invariant} (ScatNet-2) and 2-stage Contractive auto-encoders \cite{rifai2011contractive} (CAE-2) as reported in \cite{chan2015pcanet}. The parameters of \textit{HybridNet} (and other networks) are set as $L_1$ = 9, $L_2$ = 8, \textit{k}$_1$ = \textit{k}$_2$ = 7, with a block size of $B= 7 \times 7$ keeping size of overlapping regions equal to half of the block size for feature pooling. 

\item [\textbf{2}] CUReT dataset \cite{varma2009statistical}, consists of  $61$ texture categories, where each category has images of the same material with different poses, illumination, specularity, shadowing, and surface normals. Following the standard procedure in \cite{varma2009statistical,chan2015pcanet} a subset of $92$ cropped images were taken from each category and randomly partitioned into train and test sets with a split ratio of 50\%. The classification results are averaged over $10$ different trails with hyper-parameters as $L_1$ = 9, $L_2$ = 8, \textit{k}$_1$ = \textit{k}$_2$ = 5, block size $B$ = $50 \times 50$ with overlapping regions equal to half of the block size. Again, we cite the results of the baselines techniques as published in \cite{chan2015pcanet}.     

\item [\textbf{3}] ORL\footnote{The ORL database is publicly available and can be obtained from the website: http://www.uk.research.att.com/facedatabase.html} and Extended Yale-B \cite{georghiades2001few} datasets are utilized to investigate the performance for face recognition. The ORL dataset consists of 10 different images of 40 distinct subjects taken at different times, varying the lightning, facial expression, and facial details. The Extended Yale-B dataset consists of face images from 38 individuals under 9 poses and 64 illumination conditions. The images in both the datasets are cropped to size $64 \times 64$ pixels followed by unit length normalization. The classification results are averaged over $5$ different trails by progressively increasing the number of training examples\footnote{The training and test split are obtained from \url{http://www.cad.zju.edu.cn/home/dengcai/Data/FaceData.html}} with hyper-parameters as $L_1$ = 9, $L_2$ = 8, \textit{k}$_1$ = \textit{k}$_2$ = 5 with a non-overlapping block of size $B$ = $7 \times 7$.

\item [\textbf{4}] CIFAR-10 \cite{krizhevsky2009learning} dataset consists of $RGB$ images of dimensions $32 \times 32$ for object recognition consisting of $50K$ and $10K$ images for training and testing respectively. These images are distributed among $10$ classes and vary significantly in object positions, object scales, colors, and textures within each class. We varied the number of filters in layer 1 i.e., $L_1$ as 9 and 27 and kept the number of filters in layer 2 i.e. $L_2$ = 8. The patch-size \textit{k}$_1$ and \textit{k}$_2$ are kept equal and varied as 5, 7, and 9  with block size  $B$ = $8 \times 8$. Following \cite{chan2015pcanet} we also applied spatial pyramid pooling (SPP) \cite{grauman2005pyramid} to the output layer of \textit{HybridNet} (and similarly to the out layer of other networks). We additionally applied PCA to reduce the dimension of each pooled feature to 100\footnote{Results does not vary significantly on increasing the projection dimensions.}. These features are utilized with Linear-$SVM$ for classification and \textit{Attn-HybridNet} for obtaining attention features $F_{attn}$.

\item [\textbf{5}] CIFAR-100 \cite{krizhevsky2009learning} dataset closely follows CIFAR-10 dataset and consists 50k training and 10k testing images roughly distributed among 100 categories. We use the same experimental setup as in CIFAR-10 on this dataset.
\end{itemize}


\subsection{Baselines} \label{baselines}
On face recognition datasets, we compare the performance of \textit{Attn-HybridNet} and \textit{HybridNet} against three recent baselines: Deep-NMF \cite{trigeorgis2016deep}, PCANet+ \cite{low2017stacking}, and PCANet-II \cite{fan2018pcanet}\footnote{The paper did not provide its source code and the results are based on our independent implementation of their second order pooling technique.}. On MNIST variations and CuReT dataset we select the baselines as in \cite{chan2015pcanet}. Besides, the hyper-parameters of these baselines are set equal to those in \textit{HybridNet}.  

On CIFAR-10 dataset, we compare our schemes against multiple comparable baselines such as Tiled CNN \cite{ngiam2010tiled}, CUDA-Convnet \cite{alex_convnet}, VGG style CNN (VGG-CIFAR-10 reported by \cite{VGG}), K-means (tri) \cite{coates2011analysis}, Spatial Pyramid Pooling for CNN (SPP) \cite{he2015spatial}, Convolution Bag-of-Features (CBoF) \cite{passalis2017learning}, and Spatial-CBoF \cite{passalis2018training}.  Note that, we do not compare our schemes against schemes which aim to either compress deep neural networks or transfer pre-learned CNN filters such as in \cite{keshari2018learning, yu2017compressing, lin2018holistic,lin2018accelerating,he2018amc,luo2017thinet} as these schemes do not train a deep network from scratch whereas our proposed schemes and comparable baselines do so. Besides, we also report the performances of ResNet \cite{he2016deep} and DenseNet \cite{huang2017densely} on CIFAR datasets, as mentioned in their respective publication.

Lastly, we perform a qualitative case study on the CIFAR-10 dataset by studying the performance of baselines and our scheme by varying the size of the training data. Besides, we also studied the effect on the discriminability of hybrid features with attention-based fusion scheme.


\section{Results and Discussions} \label{ressec}

Our two main contributions in this research are - 1) the integration of information available from both the amalgamated view (i.e., the unique information) and the minutiae view (i.e., the common information), and 2) attention-based fusion of information obtained from these two views for supervised classification. We evaluate the significance of these contributions under the following research questions:

\textit{Q1: Is the integration of both the minutiae view and the amalgamated view beneficial? Or, does their integration deteriorate the generalization performance of {HybridNet}?}

In order to evaluate this, we varied the amount of training data in \textit{HybridNet}, the PCANet, and TFNet and obtained the classification performance from their corresponding features on CIFAR-10 and MNIST variations datasets. We cross-validated the performances of these schemes for $5$ times and present their mean and variances in Fig.~\ref{Q1}.

Firstly, these plots clearly suggest that the classification accuracies obtained with the features from \textit{HybridNet} (and also from the PCANet and the TFNet) linearly increase with respect to the size of training data. Secondly, these plots also demonstrate that the information obtained from the amalgamated view in PCANet is superior than the information obtained from the minutiae view TFNet on object-recognition dataset. However, these two kinds of information achieve competitive classification performance on variations of handwritten digits dataset which contains nearly aligned images. 

Most importantly, these plots unambiguously demonstrate that integrating both kinds of information can enhance the superiority of feature representations, consequently improving the classification performance in proposed \textit{HybridNet}. 

\textit{Q2: How does the hyper-parameters affect the discriminability of feature representations? Moreover, how does these affect the performance of \textit{HybridNet} and \textit{Attn-HybridNet}?}

\begin{table}[t]
\centering
\captionsetup{justification=centering}
\scriptsize
\setlength{\tabcolsep}{3.5pt}
\begin{tabular}{@{}cccc|ccc|c@{}}
\toprule[1pt]
\multicolumn{4}{c}{Parameters}                       & PCANet \cite{chan2015pcanet} &   TFNet  \cite{VermaL0Z18}             &    \textit{HybridNet}        &    \textit{Attn-HybridNet}              \\ \cmidrule(lr){1-8}
\multicolumn{1}{l}{$\textit{L}_1$} & ${\textit{L}_2}$ & \multicolumn{1}{l}{$\textit{k}_1$} & $\textit{k}_2$ & Error (\%) & \multicolumn{1}{c}{Error (\%)} & \multicolumn{1}{c}{Error (\%)} & \multicolumn{1}{c}{Error (\%)} \\ \cmidrule(lr){1-8} \cmidrule(lr){4-5}
8                     & 8 & 5                     & 5 & 34.80 	 & 32.57                     &  31.39        &      \textbf{28.08}      \\

8                     & 8 & 7                     & 7 & 39.92 	 & 37.19                     &            35.24     &    \textbf{30.94}      \\

8                     & 8 & 9                     & 9 & 43.91  	&        39.65            &  38.04        &   \textbf{35.33}              \\  \midrule

27                    & 8 & 5                     & 5 & 26.43  	&           29.25                &        23.84      &   \footnotesize{\textbf{18.41}}        \\
27                    & 8 & 7                     & 7 & 30.08  	& 32.57                     &  28.53       &       \textbf{25.67}       \\
27                    & 8 & 9                     & 9 & 33.94 	 &         34.79                  &    31.36  & \textbf{27.70} \\ \bottomrule[1pt]
                        
\end{tabular}
\caption{Classification Error obtained by varying hyper-parameters on CIFAR-10 dataset.}
\label{comp_attnres_param}
\end{table}

\begin{figure}[t]
\begin{center}
\captionsetup{justification=centering}
\includegraphics[width=0.31\textwidth]{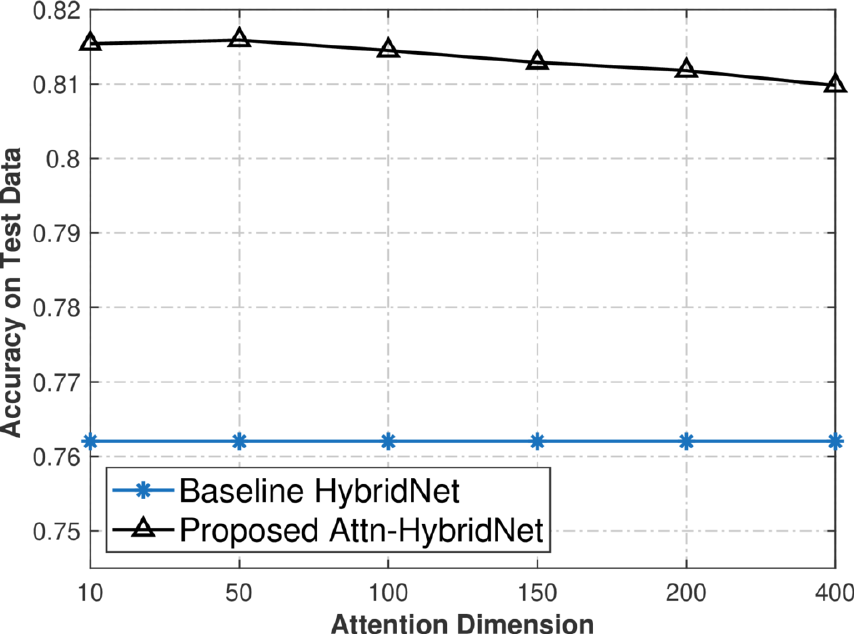}
\setlength{\belowcaptionskip}{-0.5cm}
\caption{Accuracy of Attn-HybridNet on CIFAR-10 dataset by varying the dimension of $w$ in Alg.~\ref{Algo:Attn}.}
\label{attncompf}
\end{center}
\end{figure}

To address this question, we present a detailed study on how the hyper-parameters affect the performance of \textit{HybridNet} and \textit{Attn-HybridNet}. In this regard, we compare the classification performance of the PCANet, TFNet, \textit{HybridNet}, and \textit{Attn-HybridNet} on CIFAR-10 dataset in Table~\ref{comp_attnres_param}. The lowest error is highlighted in slightly larger font, while the minimum error achieved in each row is highlighted in bold font. Moreover, we also illustrate the performance of \textit{Attn-HybridNet} by varying the dimension of context level feature vector $w$ utilized in our attention-fusion scheme in Fig.~\ref{attncompf}.

A clear trend is visible in Table~\ref{comp_attnres_param} among the performances of all the networks, where the classification error decreases with an increase in the number of filters in the first layer of the networks. This trend also demonstrates the effect of the factorization rank while obtaining the \emph{principal-components} and the matrix factors with \textit{LoMOI}; signifying that increasing the number filters in the first layer allows all the networks to increase the data variability that aids in obtaining better feature correspondences in the output stage. In addition, this also increases the dimensionality of the features extracted by the networks suggesting that comparatively higher dimensional features have lower intraclass variability among the feature representations of objects from the same category.


\setlength\tabcolsep{7pt}
\begin{table}[t]
\begin{tabular}{@{}lccc@{}}
\toprule[1pt]
\multicolumn{1}{c}{{ORL - Dataset}} & \multicolumn{3}{c}{Number of Training Instances} \\ \cmidrule(l){2-4} 
\multicolumn{1}{c}{}                               & 4                             & 6                            & 8                            \\ \midrule[1pt]
Deep-NMF \cite{trigeorgis2016deep}                                          &    9.50 $\pm$ 1.94            &     6.50 $\pm$ 2.59          &   1.75 $\pm$ 2.09     \\
PCANet-II \cite{fan2018pcanet}                                          &    16.16 $\pm$ 1.29            &     7.87 $\pm$ 1.29          &   5.00 $\pm$ 2.05     \\
PCANet+ \cite{low2017stacking}                                 &    1.25 $\pm$ 0.83            &     0.50 $\pm$ 0.52          &   0.25 $\pm$ 0.55      \\ \midrule
PCANet  \cite{chan2015pcanet}                                           &    1.75 $\pm$ 0.95            &     0.37 $\pm$ 0.34          &   0.40 $\pm$ 0.68         \\
TFNet   \cite{VermaL0Z18}                                          &    1.98 $\pm$ 0.54            &     0.50 $\pm$ 0.68          &   0.25 $\pm$ 0.59          \\
\textit{HybridNet}  (proposed)                                        &    \textbf{1.48} $\pm$ \textbf{0.72}            &     \textbf{0.25} $\pm$ \textbf{0.32}          &   \textbf{0.21} $\pm$ \textbf{0.55}           \\
\textit{Attn-HybridNet} (proposed)                                     &    5.43 $\pm$ 0.78                      &    3.11 $\pm$ 0.27                          &   1.13 $\pm$ 0.31                           \\ \bottomrule[1pt]
\end{tabular}
\caption{Classification Error on ORL dataset.}
\label{ORL}
\end{table}

\setlength\tabcolsep{2pt}
\begin{table}[]
\begin{tabular}{@{}lcccll@{}}
\toprule[1pt]
\multicolumn{1}{c}{{YaleB - Dataset}} & \multicolumn{4}{c}{Number of Training Instances} \\ \cmidrule(l){2-5} 
\multicolumn{1}{c}{}                                       & 20      & 30      & 40      & \qquad 50      \\ \midrule[1pt]
Deep-NMF \cite{trigeorgis2016deep}                           &  10.94$\pm$0.89         &    8.03$\pm$0.61      &   5.43$\pm$0.94       &  4.78$\pm$0.76      \\
PCANet-II \cite{fan2018pcanet}                           &  11.40$\pm$0.97         &    5.54$\pm$1.49      &   2.86$\pm$0.35       &  1.98$\pm$0.75      \\
PCANet+ \cite{low2017stacking}                        &  1.15$\pm$0.14       &    0.28$\pm$0.07     &    0.23$\pm$0.14     &   0.22$\pm$0.23      \\ \midrule
PCANet \cite{chan2015pcanet}                                                   &   1.35$\pm$0.17        &   0.40$\pm$0.18       &   0.38$\pm$0.16       &     0.38$\pm$0.13     \\
TFNet \cite{VermaL0Z18}                                                       & 1.97$\pm$0.27         &  0.91$\pm$0.28        &  0.40$\pm$0.16        &  0.42$\pm$0.21        \\
\textit{HybridNet} (proposed)                                                &  \textbf{1.32}$\pm$\textbf{0.35}        &  \textbf{0.55}$\pm$\textbf{0.26}        &  \textbf{0.32}$\pm$\textbf{0.25}        &   \textbf{0.34}$\pm$\textbf{0.21}       \\
\textit{Attn-HybridNet} (proposed)                                              &   5.11$\pm$0.65      &   2.80$\pm$0.42      &  2.12$\pm$0.15       &   1.88$\pm$0.40      \\ \bottomrule[1pt]
\end{tabular}
\caption{Classification Error on Extended YaleB dataset.}
\label{Yaleb}
\end{table}


\begin{table*}[t!] 
\centering
\subfloat[MNIST Variations Datasets] 
{
\begin{tabular}{@{}l|c|c|c|c|c|c|c@{}}
\toprule[1pt]
\multicolumn{1}{c}{Methods} & baisc & rot  & bg-rand & bg-img & bg-img-rot & rect-image & convex\\ \midrule
CAE-2  \cite{rifai2011contractive}                    & 2.48  & 9.66 & 10.90   & 15.50  & 45.23      & 21.54 &  - \\
TIRBM  \cite{sohn2012learning}                     & -     & \textbf{4.20} & -       & -      & 35.50      & -      &  - \\
PGBM \cite{sohn2013learning}                & -     & -    & 6.08    & 12.25  & 36.76      & \textbf{8.02} & -    \\
ScatNet-2  \cite{bruna2013invariant}                 & 1.27  & 7.48 & 12.30   & 18.40  & 50.48      & 15.94   & 6.50\\ \midrule
PCANet \cite{chan2015pcanet}                     & 1.07      & 6.88 &      6.99   &     11.16   &     35.46       & 13.59   & 4.15 \\
TFNet  \cite{VermaL0Z18}                     & 1.07  &  7.15    &   6.96      & 11.44  &   37.02         &     16.87 &  4.98  \\
\textit{HybridNet} (proposed)                      &   1.01    &  6.32    & 5.46    & 10.08  &  33.87          &   12.91   &  3.55  \\ \midrule
\textit{Attn-HybridNet} (proposed)       &   \textbf{0.94}    &  4.31    & \textbf{3.73}    & \textbf{8.68}  &  \textbf{31.33}          &   10.65   &  \textbf{2.81}  \\
\bottomrule[1pt]
\end{tabular}
}
\quad
\subfloat[CuReT Dataset]
{
\begin{tabular}{@{}l|c@{}}
\toprule[1pt]
Methods & Error (\%)\\ \midrule
Textons \cite{hayman2004significance} & 1.50        \\
BIF \cite{crosier2010using}     & 1.40        \\
Histogram \cite{broadhurst2005statistical}     & 1.00        \\
ScatNet \cite{bruna2013invariant} & \textbf{0.20}        \\  \midrule
PCANet \cite{chan2015pcanet}  &     0.84        \\
TFNet \cite{VermaL0Z18}   &   0.96          \\
\textit{HybridNet} (proposed)   &      0.81      \\ \midrule
\textit{Attn-HybridNet} (proposed) &  0.72 \\ \bottomrule[1pt]
\end{tabular}
}
\caption{Classification Error on MNIST variations and CUReT datasets.}
\label{mnistvar}
\end{table*}

\setlength\tabcolsep{10pt}
\begin{table}[t]
\centering
\captionsetup{justification=justified}
\begin{tabular}{@{}lccl@{}}
\toprule[1pt]
\textbf{Methods} &  \multicolumn{1}{c}{\textbf{\#Depth}} & \multicolumn{1}{c}{\textbf{\#Params}} & \multicolumn{1}{c}{\textbf{Error}} \\ \midrule
Tiled CNN \cite{ngiam2010tiled} & - & - & 26.90        \\
K-means (tri.) \cite{coates2011analysis} (1600 dim.) & 1 & 5 & 22.10        \\
CUDA-Convnet \cite{alex_convnet} & 4 & 1.06M &  18.00   \\
VGG-CIFAR-10 \cite{Ilia} & 5 & 2.07M  &   20.04 \\ 
SPP  \cite{he2015spatial}                 & 5  & 256.5K & 19.39 \\
CBoF  \cite{passalis2017learning}                 & 5  & 174.6K & 20.47 \\
Spatial-CBoF  \cite{passalis2018training}         & 5  & 199.1K & 21.37 \\
ResNet reported in-\cite{huang2016deep}     & 110  & 1.7M & 13.63 \\
DenseNet-BC reported in-\cite{huang2017densely}     & 250  & 15.3M & \textbf{5.2} \\\midrule
PCANet \cite{chan2015pcanet} & 3 & 7 &    26.43        \\
TFNet \cite{VermaL0Z18}   & 3 & 7  & 29.25          \\
\textit{HybridNet} (proposed)   & 3 &  7 &   23.84      \\
\textit{Attn-HybridNet} (proposed)  & 3 & 12.7k &  18.41      \\ \bottomrule[1pt]
\end{tabular}
\caption{Classification Error on CIFAR-10 dataset with no data augmentation. The DenseNet achieves the lowest classification error but at the expense of huge depth and substantial computational cost among all techniques.}
\label{Q3res_c}
\vspace{-0.2cm}
\end{table}

Another trend can be observed in the performance table where the classification error increases with the increase of the patch size of the image. Since the dimension of images in CIFAR-10 is $32 \times 32$, this may be due to the presence of less background with smaller image-patches as increasing the patch size gradually mount to non-stationary data \cite{chan2015pcanet}.    

Importantly, our proposed \textit{Attn-HybridNet} substantially reduces the classification error by \textbf{22.78}$\%$ when compared to classification performance with \textit{HybridNet} on CIFAR-10 dataset. The plot in Fig.~\ref{attncompf} shows the effect on classification accuracy by varying dimensions of feature level context vector $w$ in \textit{Attn-HybridNet}.

\textit{Q3: How does the proposed \textit{Attn-HybridNet} (and \textit{HybridNet}) perform in comparison to the baseline techniques?}

To evaluate this requirement, we compare the performance of the proposed \textit{Attn-HybridNet} and \textit{HybridNet} against baselines as detailed in Sec.~\ref{baselines}. In this regard, we present the performance comparison on face recognition datasets in Table~\ref{ORL} and Table~\ref{Yaleb}.The performance comparison on handwritten digits and texture classification datasets are presented in Table~\ref{mnistvar}. Furthermore, classification results on CIFAR-10 and CIFAR-100 datasets are presented in Table~\ref{Q3res_c} and Table~\ref{Q3res_c2}, respectively.

\begin{figure}[t]
\begin{center}
\captionsetup{justification=centering}
\includegraphics[width=0.31\textwidth]{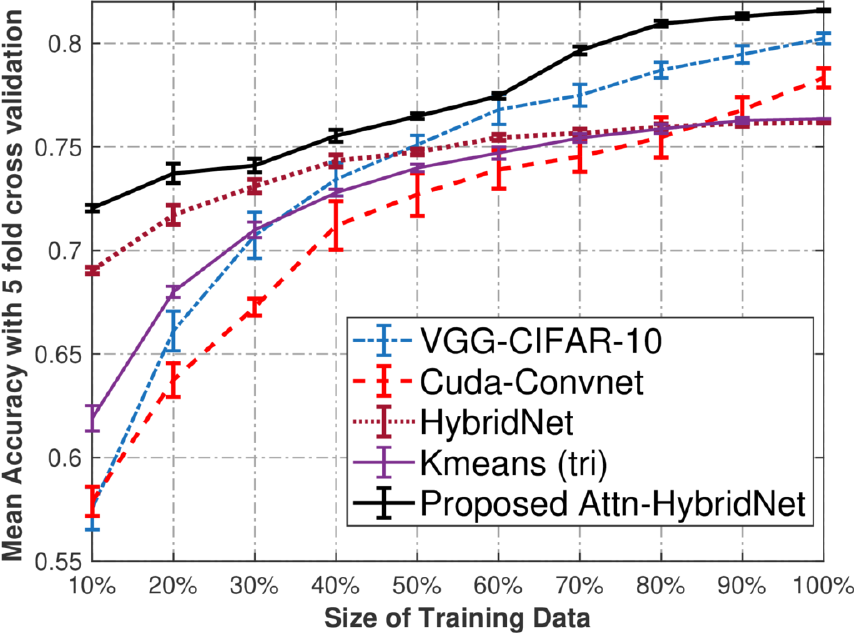}
\setlength{\belowcaptionskip}{-0.2cm}
\caption{(Best viewed in color) Accuracy of various methods on CIFAR-10 dataset by varying size of the training data}
\label{attncomp}
\end{center}
\end{figure}

\begin{figure}[t]
    \centering
    \captionsetup{justification=centerlast, skip=5pt}
        \subfloat[\textit{HybridNet}]{\includegraphics[width=0.22\textwidth]{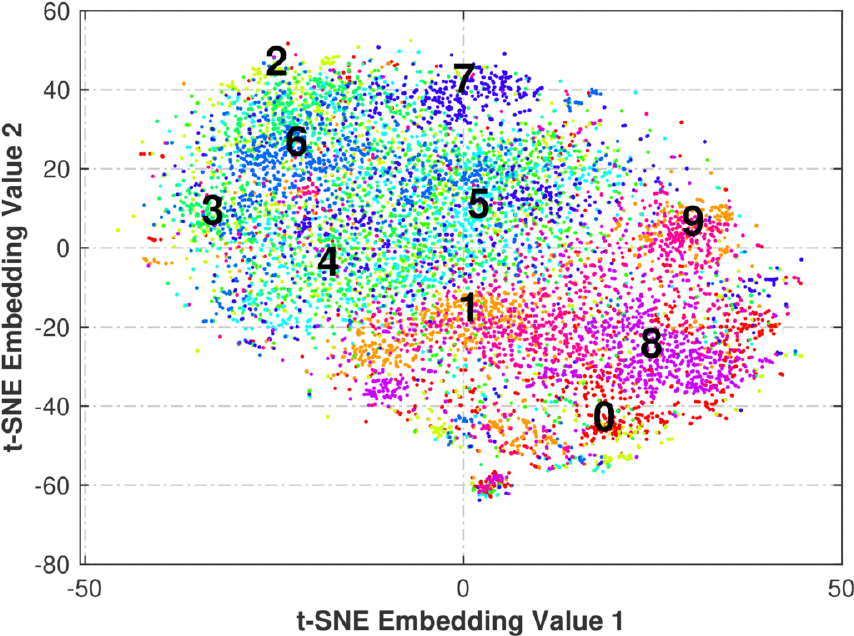}}
        \quad
        \subfloat[\textit{Attn-HybridNet}]{\includegraphics[width=0.22\textwidth]{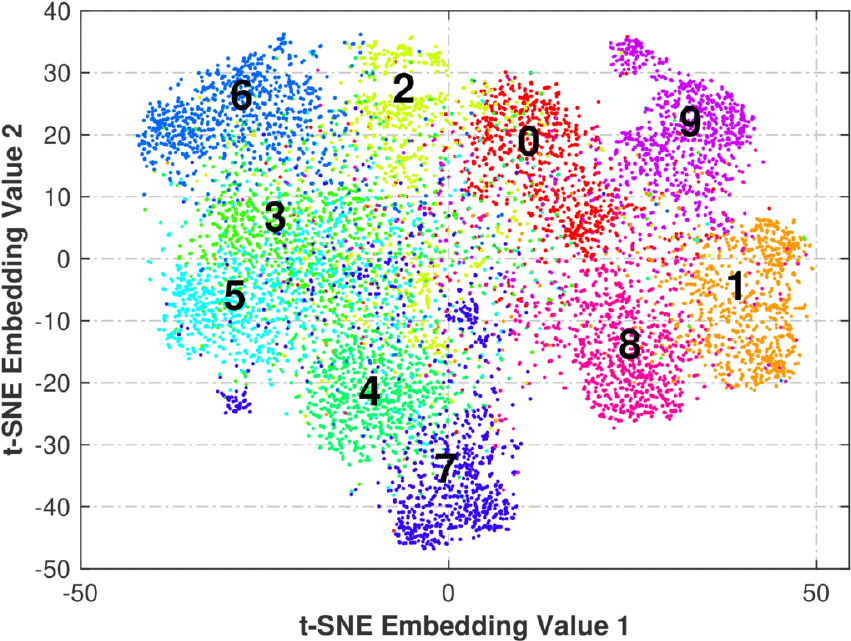}}
        \setlength{\belowcaptionskip}{-0.2cm}
        \caption{(Best viewed in color) t-SNE visualization of features from \textit{HybridNet} and \textit{Attn-HybridNet} on CIFAR-10 dataset.}
\label{tSNE}
\end{figure}

Besides, we visualize the discriminability of feature representation obtained from \textit{HybridNet} and \textit{Attn-HybridNet} with t-SNE plot \cite{maaten2008visualizing} in Fig.~\ref{tSNE} to perform a qualitative analysis of their discriminability. We further aid this analysis with a plot to study their classification performance obtained with increasing amount of training data on CIFAR-10 dataset in Fig.~\ref{attncomp}.

\paragraph{Performance on face recognition} A similar trend is noticeable from the classification performances on ORL and Extended YaleB datasets. First, for all schemes, the classification error decreases with the increase of the number of training examples. This is expected as by increasing the amount of training data all schemes can better estimate the variation in lighting, facial expression, and pose. Secondly, among the baselines, PCANet+ performs substantially better than Deep-NMF and PCANet-II on both the face datasets. The poor performance for Deep-NMF can be justified as it needs a large amout of data to estimate its parameters. Whereas for PCANet-II, the explicit alignment of face images as a requirement can explain its degradation in performance. Besides, the PCANet+ also performs slightly better than PCANet as the earlier enhances the latter with a better feature encoding scheme. 

Lastly, our proposed \textit{HybridNet} outperforms the baselines and individual schemes, i.e. PCANet and TFNet, on both the datasets. This validates our hypothesis that the common and unique information are both essential and their fusion can enhance the classification performance. However, the classification performance of \textit{Attn-HybridNet} is slightly worse than \textit{HybridNet}. Again, this might be due to less amount of data available while learning the attention parameters, similar to Deep-NMF albeit it still performs better than Deep-NMF as; the underlying features are highly discriminative, and therefore it is less strenuous to discover attention weights for the proposed fusion scheme in comparison to Deep-NMF.
 
\paragraph{Performance on digit recognition and texture classification}On MNIST handwritten digits variations dataset, the \textit{Attn-HybridNet} (and also the \textit{HybridNet}) outperforms the baselines on five out of seven variations. In particular, for \emph{bg-rand} and \emph{bg-img} variations, we decreased the error (compared to \cite{VermaL0Z18}) by \textbf{31.68}$\%$ and \textbf{13.80}$\%$ respectively. On CUReT texture classification dataset, the \textit{Attn-HybridNet} achieves the lowest classification error among all the networks, albeit it achieves slightly higher classification error compared to state of the art. However, the difference in classification error achieved by state of the art \cite{bruna2013invariant} and \textit{Attn-HybridNet} is marginal and is only $0.5\%$. 

\setlength\tabcolsep{10pt}
\begin{table}[t]
\centering
\captionsetup{justification=justified}
\begin{tabular}{@{}lccl@{}}
\toprule[1pt]
\textbf{Methods} &  \multicolumn{1}{c}{\textbf{\#Depth}} & \multicolumn{1}{c}{\textbf{\#Params}} & \multicolumn{1}{c}{\textbf{Error}} \\ \midrule
ResNet reported in-\cite{huang2016deep}     & 110  & 1.7M & 37.80 \\
DenseNet-BC reported in-\cite{huang2017densely}     & 250  & 15.3M & \textbf{19.64} \\\midrule
PCANet \cite{chan2015pcanet} & 3 & 7 &    53.00        \\
TFNet \cite{VermaL0Z18}   & 3 & 7  & 53.63          \\
\textit{HybridNet} (proposed)  & 3 &  7 &   49.87      \\
\textit{Attn-HybridNet} (proposed)  & 4 & 42.2k &  47.44    \\ \bottomrule[1pt]
\end{tabular}
\caption{Classification Error on CIFAR-100 dataset with no data augmentation.}
\label{Q3res_c2}
\vspace{-0.2cm}
\end{table}

\paragraph{Quantitative Performance on CIFAR Datasets}
We compare the performance of proposed \textit{Attn-HybridNet} and \textit{HybridNet} against baselines as described in Sec.~\ref{baselines} on CIFAR-10 and CIFAR-100 datasets and report their respective accuracies in Table~\ref{Q3res_c} and Table~\ref{Q3res_c2} respectively.

The proposed \textit{Attn-HybridNet} achieves the best performance among all kinds of networks studied in the paper. Technically, the proposed \textit{HybridNet} reduces the error by $9.80\%$ on CIFAR-10 and $5.91\%$ on CIFAR-100 dataset in comparison to min(PCANet, \textit{HybridNet}). Additionally, the proposed \textit{Attn-HybridNet} further reduces the error by $22.78\%$ on CIFAR-10 and $4.87\%$ on CIFAR-100 dataset in comparison to \textit{HybridNet}.  
Besides, on CIFAR-10 dataset, \textit{Attn-HybridNet} achieves substantially lower error compared to Titled CNN \cite{ngiam2010tiled}, K-means (tri), and the PCANet; particularly 16.70$\%$ lower than K-means (tri) which has $2 \times$ higher feature dimensionality than our proposed \textit{HybridNet} and utilizes $L_2$ regularized-$SVM$ instead of $Linear$-$SVM$ for classification.

The performance of our proposed \textit{Attn-HybridNet} is still better than VGG-CIFAR-10 \cite{VGG} and comparable to CUDA-Convnet  \cite{alex_convnet}\footnote{We cite the accuracy as published.}, both of which have more depth than the proposed \textit{Attn-HybridNet}. In particular, we have reduced the error by $1.63\%$ than VGG-CIFAR-10 with $99.63\%$ less trainable parameters. At the same time, we have performed very competitive to CUDA-Convnet achieving $0.41\%$ higher error rate but with $88\%$ less number of tunable parameters. 

Our proposed \textit{Attn-HybridNet} also performs marginally better in compared to deep-quantized networks such as SPP \cite{he2015spatial}, CBoF \cite{passalis2017learning}, and Spatial-CBoF \cite{passalis2018training}. This is because the proposed scheme is decoupled as feature extraction and feature pooling schemes and hence the effort required to estimate the tuneable parameters is negligible. The quantization schemes are proposed to reduce the parameters in the fully connected layer but requiring the same efforts required to find optimal parameters of the higher layers. 

Besides, the performance gap between \textit{Attn-HybridNet} and state of the art ResNet and DenseNet are not comparable as the depth and the computational complexity of the latter networks are tremendously huge. Therefore, the main bottleneck for these schemes is requirement of high performing hardware which is opposite to the motivation of this work that is alleviation of such requirementsand hence the tradeoff. 

\paragraph{Qualitative Discussion on CIFAR-10} \label{qcase}
We now present a qualitative discussion on the performances of various baselines and our proposals by varying the size of training data in Fig.~\ref{attncomp}. Although our proposed \textit{Attn-HybridNet} consistently achieved the highest classification performance, a few interesting patterns are noticeable in the performance curves.

A paramount observation in this regard is the lower classification performance achieved by both CUDA-Convnet \cite{alex_convnet} and VGG-CIFAR-10 \cite{Ilia} with less amount of training dataset, particularly until 40$\%$. It is intuitive and justifiable since less amount of the training data is not sufficient to learn the parameters of these deep networks. However, on increasing the amount of training data (above 50$\%$), the performance of these networks increases substantially i.e., increases with a larger margin compared to the performance of SVM based schemes in \textit{HybridNet} and K-means (tri) \cite{coates2011analysis}.  

The second observation is regarding the classification performances of \textit{HybridNet} and K-means (tri). Both these networks achieve higher classification accuracy compared to the deep networks with less amount of training data; particularly, the \textit{HybridNet} has \textbf{11.56}$\%$ higher classification rate compared to the second-highest classification accuracy achieved by K-means (tri) with only $10\%$ of the training dataset. However, the accuracy of these networks does not scale or increase substantially with an increase in the training data, as noticed by deep-network-based schemes.

Besides, the \textit{Attn-HybridNet} achieved the highest classification performance across different sizes of the training dataset among all techniques. A possible explanation for this is the requirement of fewer parameters with proposed attention-fusion while performing feature selection with attention-based fusion to alleviate the feature redundancy.
Moreover, the t-SNE plot in Fig.~\ref{tSNE} compares the discriminability of features obtained with the \textit{HybridNet} and \textit{Attn-HybridNet}. The plot on the features obtained from \textit{Attn-HybridNet} Fig.~\ref{tSNE}(b) visually achieves better clustering than the plot on features obtained from \textit{HybridNet} Fig.~\ref{tSNE}(a) and justifies the performance improvement with our proposal.  

\section{Conclusion and Future Work}
\label{sec:con}

In this paper, we have introduced \textit{HybridNet}, which integrates the information discovery and feature extraction procedure from the amalgamated view and the minutiae view of the data. The development of \textit{HybridNet} is motivated by the fact that information obtained from the two views of the data are individually insufficient but necessary for classification. To extract features from the minutiae view of the data, we proposed the TFNet that obtains weights of its convolution-tensor filters by utilizing our custom-built \textit{LoMOI} factorization algorithm. We then demonstrated how the information obtained with the two views of data are complementary to each other. Then, we provided details to simultaneously extract the common information from the amalgamated view and unique information with the minutiae view of the data in our proposed \textit{HybridNet}. The significance of integrating these two kinds of information with \textit{HybridNet} is demonstrated by performing classification on multiple real-world datasets. 
 
Although the \textit{HybridNet} achieves higher classification accuracy, it still suffers from the problem of feature redundancy arising from the generalized spatial pooling operation utilized to aggregate the features in the output layer. Therefore, we proposed \textit{Attn-HybridNet} for alleviating the feature redundancy by performing attentive feature selection. Our proposed \textit{Attn-HybridNet} enhances the discriminability of features, which further enhances their classification performance.

We performed comprehensive experiments on multiple real-world datasets to validate the significance of our proposed \textit{Attn-HybridNet} and \textit{HybridNet}. The features extracted using our proposed \textit{Attn-HybridNet} achieved similar classification performance among popular baseline methods with significantly less amount of hyper-parameters and training time required for their optimization. Besides, we also conducted multiple case studies with other popular baseline methods to provide qualitative justifications for the superiority of features extracted by our proposed \textit{Attn-HybridNet}.

Furthermore, our research can be further improved with two interesting research directions. The first direction is the design of \textit{HybridNet} filters to accommodate various non-linearities in the data such as alignments and occlusion. A second research direction can be the design of attention-based fusion for generalized tasks such as face verification and gait recognition.  

\bibliographystyle{IEEEtran}
\bibliography{TCYB_R2}
%
\vspace{-0.8cm}
\begin{IEEEbiography}[{\includegraphics[width=1in,height=1.25in,clip,keepaspectratio]{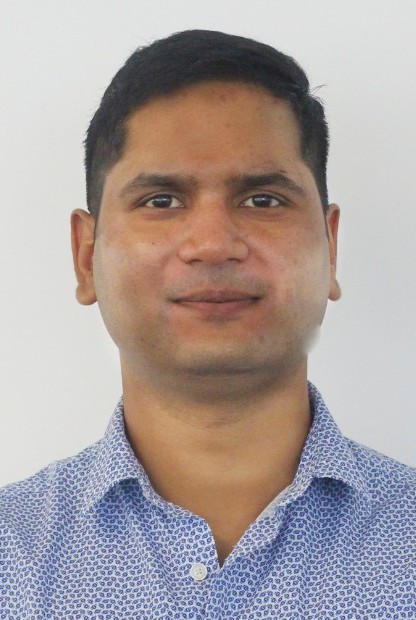}}]{Sunny Verma}
Sunny Verma received his Ph.D. degree in Computer Science from University of Technology Sydney in 2020. He is currently working as Postdoctoral Research Fellow at the Data Science Institute, University of Technology Sydney, and as a visiting scientist at Data61, CSIRO. Before joining UTS, he was a Research Assistant at Department of Electrical Engineering, IITD India, and then worked as Senior Research Assistant at Hong Kong Baptist University, Hong Kong. He obtained his Ph.D. from the University of Sydney. His research interests include data mining, fairness in machine learning, interpretable deep learning, and open-set recognition systems.
\end{IEEEbiography}

\vspace{-0.8cm}
\begin{IEEEbiography}[{\includegraphics[width=1in,height=1.25in,clip,keepaspectratio]{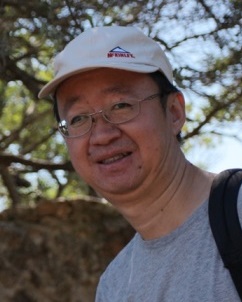}}]{Chen Wang}
Chen Wang is a senior research scientist with Data61, CSIRO. His research is in distributed and parallel computing with recent focus on data analytics systems and deep learning interpretability. He published more than 70 papers in major journals and conferences such as TPDS, TC, WWW, SIGMOD and HPDC. He has industrial experience. He developed a high-throughput event system and a medical image archive system used by many hospitals and medical centers in the USA.
\end{IEEEbiography}
\vspace{-0.8cm}
\begin{IEEEbiography}[{\includegraphics[width=1in,height=1.25in,clip,keepaspectratio]{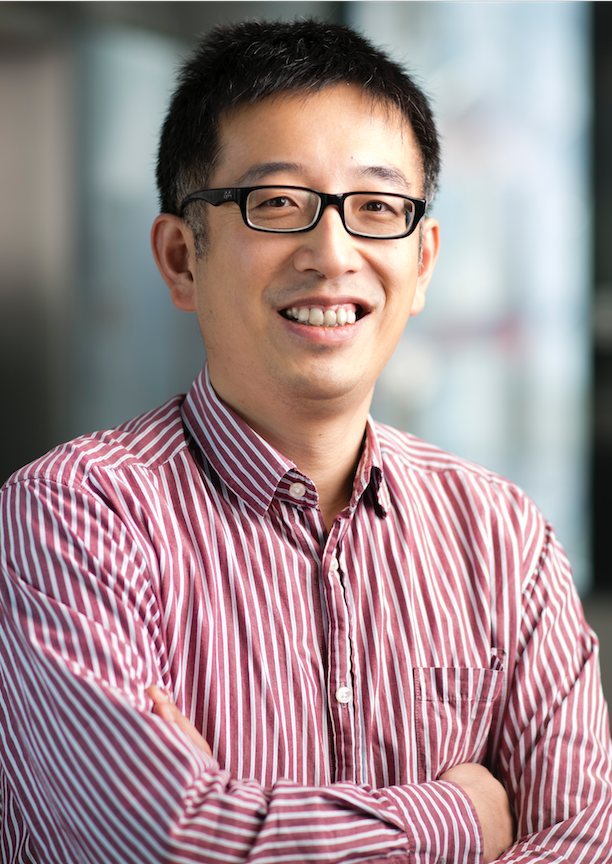}}]{Liming Zhu}
Dr/Prof. Liming Zhu is a Research Director at Data61, CSIRO. He is also a conjoint full professor at University of New South Wales (UNSW). He is the chairperson of Standards Australia's blockchain and distributed ledger committee. His research program has more than 300 people innovating in the area of big data platforms, computational science, blockchain, regulation technology, privacy and cybersecurity. He has published more than 200 academic papers on software architecture, secure systems and data analytics infrastructure and blockchain.
\end{IEEEbiography}

\vspace{-0.8cm}
\begin{IEEEbiography}[{\includegraphics[width=1in,height=1.25in,clip,keepaspectratio]{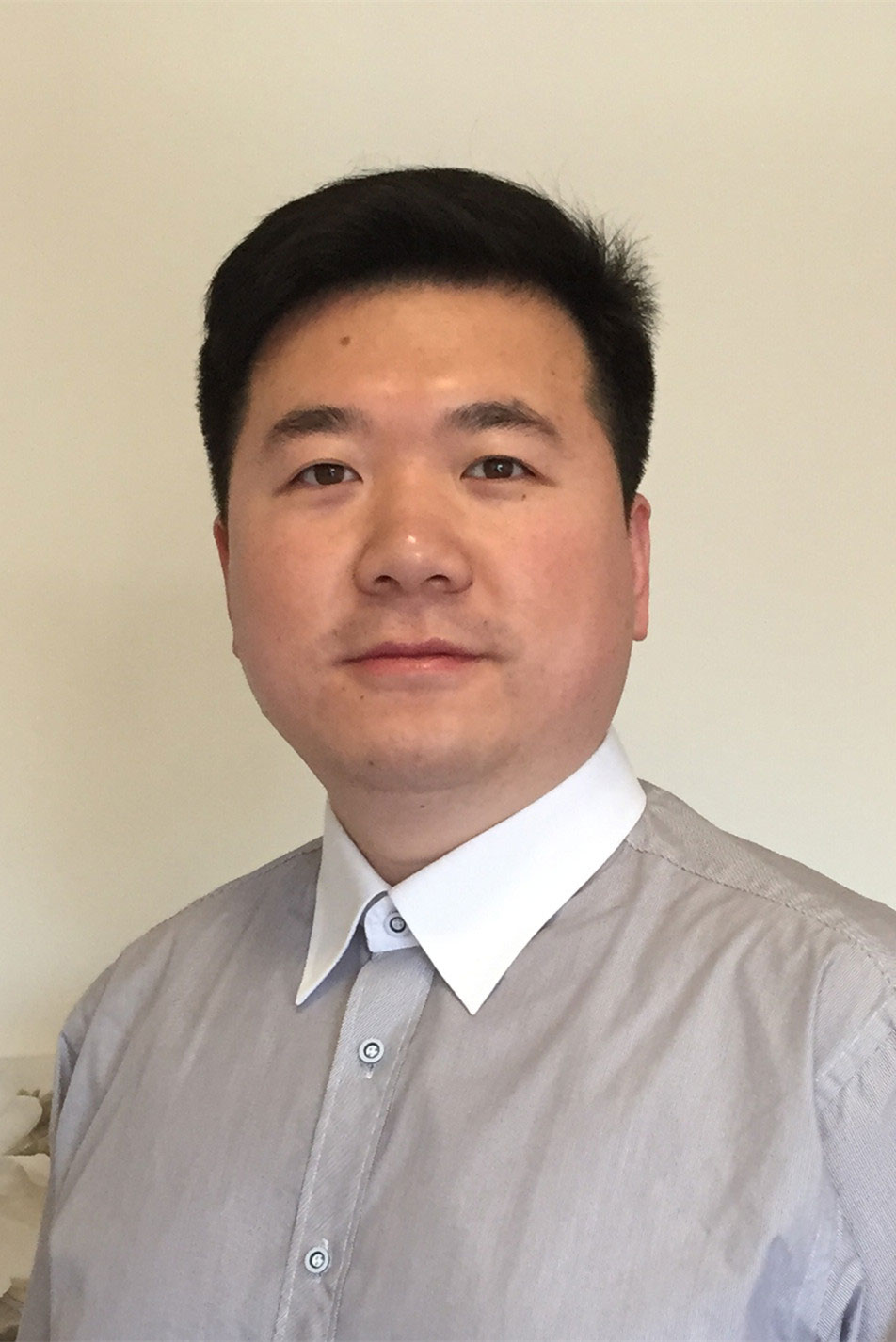}}]{Wei Liu}
Wei Liu (M'15-SM'-20) is a Senior Lecturer and the Data Science Research Leader at the Advanced Analytics Institute, School of Computer Science, University of Technology Sydney. Before joining UTS, he was a Research Fellow at the University of Melbourne and then a Machine Learning Researcher at NICTA. He obtained his PhD from the University of Sydney. He works in the areas of machine learning and data mining and has published more than 80 papers in research topics of tensor factorization, game theory, adversarial learning, graph mining, causal inference, and anomaly detection. He has won three best paper awards.
\end{IEEEbiography}

\end{document}